%% file: main.tex
\documentclass[10pt,twocolumn,letterpaper]{article}

\usepackage[pagenumbers]{cvpr} 

\input{preamble}
\usepackage{subcaption}
\usepackage[table]{xcolor}
\usepackage{colortbl}
\definecolor{lightgray}{gray}{0.93}
\definecolor{cvprblue}{rgb}{0.21,0.49,0.74}
\usepackage[pagebackref,breaklinks,colorlinks,allcolors=cvprblue]{hyperref}

\def\paperID{22596} 
\def\confName{CVPR}
\def\confYear{2026}

\author{First Author\\
Institution1\\
Institution1 address\\
{\tt\small firstauthor@i1.org}
\and
Second Author\\
Institution2\\
First line of institution2 address\\
{\tt\small secondauthor@i2.org}
}

\usepackage{subcaption} 
 
\usepackage{algorithm}
 
\usepackage{amsmath, amssymb}
\usepackage{ifthen} 
\usepackage{hyperref}
\usepackage{url}
 \usepackage{booktabs}
\usepackage{algpseudocode}
\usepackage[dvipsnames]{xcolor}

\usepackage{paracol}         
\usepackage{wrapfig}
\usepackage[most]{tcolorbox} 
\usepackage{subcaption}
\usepackage{enumitem}
 
\usepackage{amsthm}     
\usepackage{bm}         
\usepackage{mathtools}  
\usepackage{xcolor}     

\tcbset{
  findingstyle/.style={
    enhanced,
    breakable,
    sharp corners,
    boxrule=0pt,
    colback=#1!8,            
    colframe=#1!70!black,    
    borderline west={3mm}{0pt}{#1}, 
    title style={colback=#1!90!black, coltext=white},
    coltitle=white,
    fonttitle=\bfseries,
    top=4pt, bottom=3pt, left=7pt, right=4pt
  }
}
\newcommand{\lin}{\mathrm{lin}}
\newcommand{\snrf}{\mathrm{snrf}}
\newcommand{\tgt}{\mathrm{tgt}}
\newcommand{\src}{\mathrm{src}}

\newcommand{\za}[1]{{\color{black}{#1}}}

\usepackage[most]{tcolorbox}
\usepackage{xcolor}

\newcounter{finding}

\tcbset{
  findingstyle/.style={
    enhanced,
    colback=#1!5,
    colframe=#1!90!black,
    boxrule=0.8pt,
    arc=2mm,
    fonttitle=\bfseries,
  }
}

\newenvironment{finding}[2][Fuchsia]
{%
  \refstepcounter{finding}%
  \begin{tcolorbox}[findingstyle=#1,
    title={Finding~\thefinding\ #2}]
}
{%
  \end{tcolorbox}
}

\title{Do LLMs and VLMs Share Neurons for Inference? Evidence and Mechanisms of Cross-Modal Transfer}

\author{
Chenhang Cui$^{1}$ \quad
An Zhang$^{2}$\thanks{Corresponding author.} \quad
Yuxin Chen$^{1}$ \quad
Gelei Deng$^{3}$  \quad 
Jingnan Zheng$^{1}$ \\[2pt]
Zhenkai Liang$^{1}$ \quad
Xiang Wang$^{2}$ \quad
Tat-Seng Chua$^{1}$ \\[6pt]
$^{1}$National University of Singapore, Singapore \\
$^{2}$University of Science and Technology of China, Hefei, China \\
$^{3}$Nanyang Technological University, Singapore
}
\begin{document}

\maketitle
\begin{abstract}
Large vision-language models (LVLMs) have \za{rapidly advanced} across various domains, yet they still lag behind strong text-only large language models (LLMs) on tasks that require multi-step inference and compositional decision-making. 
Motivated by their shared transformer architectures, we investigate whether the two model families \za{rely on} common internal computation for such inference. 
At the neuron level, we \za{uncover a surprisingly large overlap: more than half of the top-activated units during multi-step inference are shared between representative LLMs and LVLMs, revealing a modality-invariant inference subspace.}
\za{Through causal probing via activation amplification, we further show that these shared neurons encode consistent and interpretable concept-level effects, demonstrating their functional contribution to inference.}
Building on this insight, we propose \textbf{Shared Neuron Low-Rank Fusion (SNRF)}, a parameter-efficient framework that transfers \za{mature} inference circuitry from LLMs to LVLMs. 
\textsc{SNRF} profiles cross-model activations to identify shared neurons, computes a low-rank approximation of inter-model weight differences, \za{and injects these updates selectively within the shared-neuron subspace}.
\za{This mechanism strengthens multimodal inference performance with minimal parameter changes and requires no large-scale multimodal fine-tuning}.
Across diverse \za{mathematics and perception} benchmarks, \textsc{SNRF} consistently enhances \za{LVLM} inference performance while preserving perceptual capabilities.
\za{Our results demonstrate that shared neurons form an interpretable bridge between LLMs and LVLMs, enabling low-cost transfer of inference ability into multimodal models. Our code is available at \url{https://github.com/chenhangcuisg-code/Do-LLMs-VLMs-Share-Neurons}.}
\end{abstract}
\section{Introduction}


Large vision-language models (LVLMs) have \za{achieved} rapid progress on various tasks such as captioning~\citep{blip2}, grounding~\citep{Kosmos2}, recommendation \citep{bundle}, and interaction~\citep{Rt-2}, yet—as \za{repeatedly observed in recent studies}~\citep{mmmupro,crossbench,emma}—\za{their general multi-step inference ability remains substantially weaker than that of strong text-only large language models (LLMs)}.
A prevailing approach to \za{close} this gap has been to scale LVLMs with more \za{multimodal} data and larger size~\citep{Palix,qwen2_5vl,internvl2}.
However, scaling alone faces persistent obstacles: the amount of paired image-rationale data is still limited compared to the vast text corpora~\citep{flamingo,Mint,VLCOT}, and it remains unclear how general abilities acquired by LLMs should transfer to multimodal settings. 
These limitations raise a natural question:
\textbf{Can we leverage the relatively mature capabilities of LLMs to enhance LVLMs at a relatively low cost?}

Prior \za{studies} show that models \za{sharing} similar architectures often exhibit consistent \za{internal} behavior \citep{sim_word,llm_circuit,multiling}. 
Motivated by this, we \za{investigate whether} LLMs and LVLMs display comparable consistency in inference.
\za{We perform neuron-level activation profiling \citep{safe_neuron,multiling} to identify the units that activate most strongly during inference tasks.}
Across distinct LLM and LVLM families (e.g., Qwen2.5-Math \citep{qwen2_5_math} vs. Qwen2.5-VL \citep{qwen2_5vl}; Idefics3 \citep{idefics3} vs. Deepseek-LLaMA3 \citep{deepseekr1}), we find that more than half of the top-activated neurons overlap. 
This nontrivial consistency suggests that, despite differences in modality and training corpora, these models \za{surprisingly} converge on similar internal pathways for inference.
To further probe this phenomenon, we conduct an amplification-based analysis: boosting the activations of the overlapped neurons and measuring the resulting changes in model outputs. 
This reveals \za{that the shared neurons encode coherent, functional concepts—many of them math-related—and that amplifying these units systematically strengthens inference behaviors.
These findings clearly indicate that LLMs and LVLMs share a transferable neuron subspace.}
Building on this observation, leveraging these neurons provides an efficient pathway for transferring relatively mature inference capabilities of LLMs into multimodal models.

\begin{figure}
    \centering
    \includegraphics[width=\linewidth]{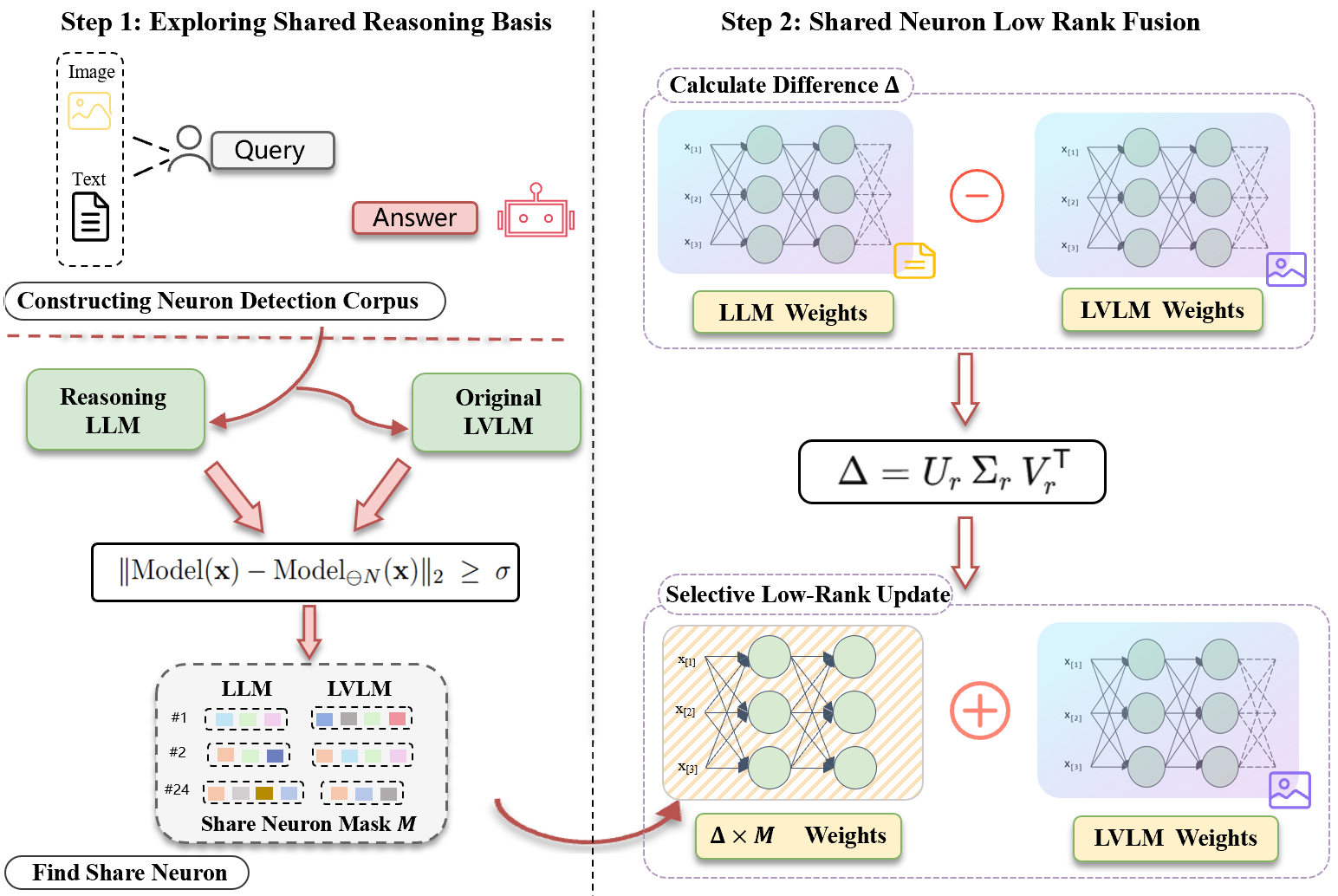}
    \caption{Overview of the proposed Shared Neuron Low-Rank Fusion (SNRF) framework.
SNRF first performs neuron-level activation profiling to identify neurons that are consistently co-activated during inference across an LLM and an LVLM.
The intersection of these units forms a shared neuron subspace, which is reused and amplified through low-rank adapters.
}
    \label{fig:framework}
\end{figure}

\begin{figure}[t]
    \centering
    \begin{subfigure}{0.48\linewidth}
        \centering
        \includegraphics[width=\linewidth]{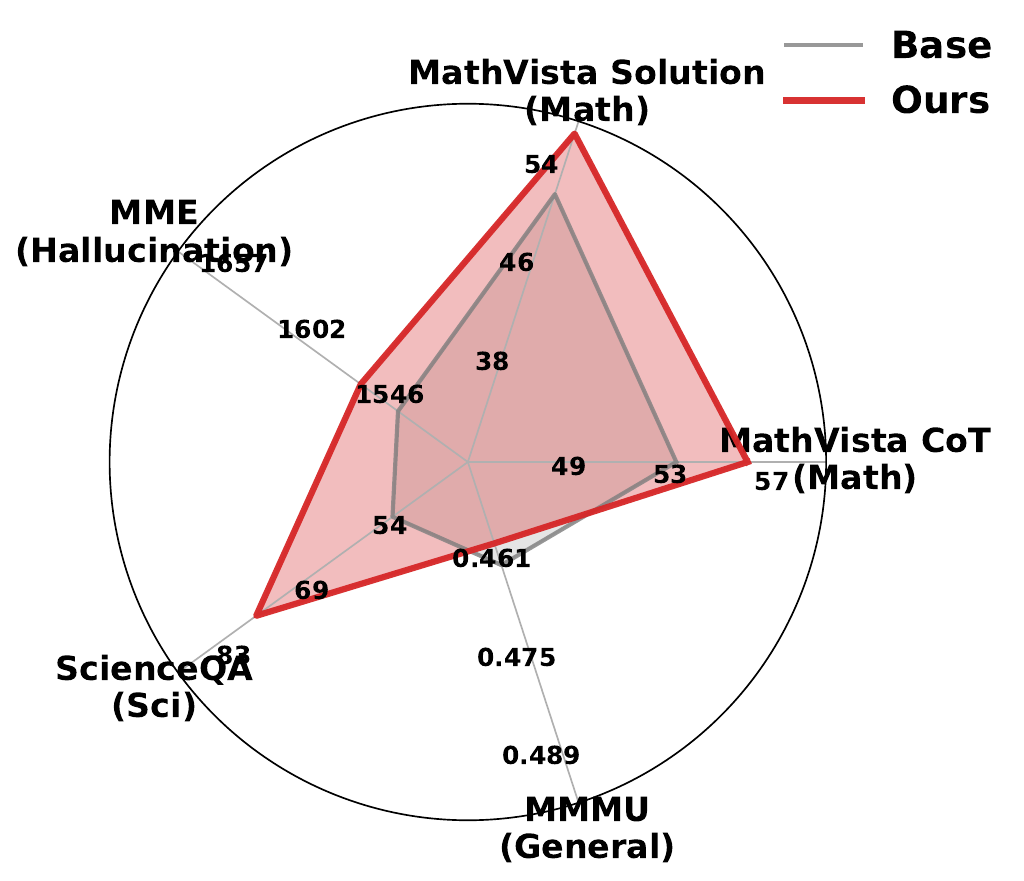}
        \caption{Qwen2.5-VL-3B}
    \end{subfigure}
    \hfill
    \begin{subfigure}{0.48\linewidth}
        \centering
        \includegraphics[width=\linewidth]{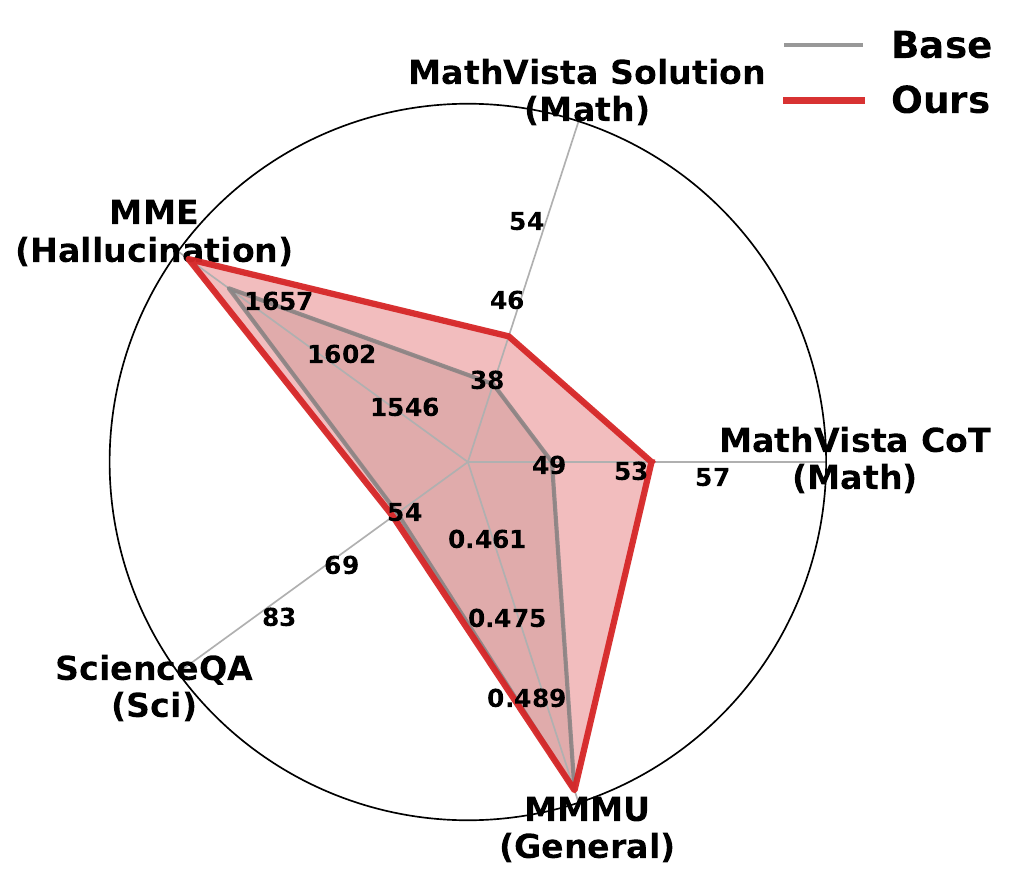}
        \caption{Qwen2.5-VL-7B}
    \end{subfigure}
    \caption{
        Comparison of Base and our method across five  multimodal  benchmarks. 
        Our method consistently enhances mathematical  performance while maintaining comparable robustness on general metrics.
    }
    \label{fig:qwen_radar}
\end{figure}

Motivated by this insight, we propose \textbf{Shared Neuron Low-Rank Fusion (SNRF)}, a parameter-efficient framework that transfers inference circuitry \za{from} an LLM to an LVLM by explicitly reusing the subset of neurons that both models activate during inference. 
The overall framework is shown in Figure \ref{fig:framework}.
\za{\textsc{SNRF} first identifies a set of \emph{shared neurons} by profiling cross-model activations on inference prompts and then computes a low-rank approximation of the inter-model weight difference.
The low-rank update is projected onto the shared-neuron subspace, injecting compact inference signals while preserving the LVLM's perceptual pathways.
\textsc{SNRF} enables efficient transfer without large-scale multimodal training.}
\textsc{SNRF} \za{is extensively evaluated} on diverse multimodal benchmarks spanning science QA, mathematics, and perception tasks. 
As shown in Figure \ref{fig:qwen_radar}, \textsc{SNRF} consistently improves LVLM performance \za{while maintaining its original visual and perceptual capabilities}.

Our contributions are threefold. First, we provide an empirical finding that LLMs and LVLMs exhibit substantial overlap among their top-activated neurons during language inference, suggesting the existence of a shared neuron subspace across modalities. Second, we introduce an amplification-based neuron analysis that enhances specific activations to reveal their functional concepts and causal influence on model outputs, enabling concept-level interpretation of inference units. Third, we propose \textsc{SNRF}, a parameter-efficient framework that aligns and reuses consistently activated neurons to transfer mature text inference into LVLMs with minimal additional cost.

\section{Related Work}
\textbf{Large Vision and Language Models.}
\za{Modern vision-language models increasingly adopt a pretrained LLM as the text decoder, paired with a vision encoder and a lightweight projector. 
This shift—from contrastive encoder-decoder frameworks (e.g., CLIP-style contrastive pretraining \citep{CONVIRT,CLIP}) to LLM-centric designs—has enabled stronger transfer and significantly easier scaling.}
Recent models (e.g., Qwen2.5-VL \citep{qwen2_5vl}, LLaVA-OneVision \citep{llavaone}) broaden inputs from single images to multi-image \za{sequences}, while a unified \za{architecture} and tighter cross-modal attention further improve fusion \za{quality}.
Beyond visual instruction tuning \citep{llava,Minigpt4}, multimodal alignment increasingly adopts reinforcement learning (RL)-based preference optimization \citep{kimivl,wang2024mpo} and verifiable-reward paradigms \citep{genreward,skyworkrmvl} to enhance stepwise fidelity, safety, and calibration. 
For inference, chain-of-thought has been extended to incorporate visual grounding \citep{multimodalcot,visualcot}, and tool-augmented pipelines  \citep{Agentthink} dispatch subproblems to external vision modules or generated code to solve complex queries.   
Despite \za{these} recent architectural and alignment advances, LVLMs still lag behind strong text-only LLMs in compositional inference under visual contexts. 
Benchmarks \za{such as} Exams-V \citep{examsv} and MMMU-Pro \citep{mmmupro} consistently reveal substantial gaps in faithfulness, robustness, and numerical accuracy, \za{suggesting that scaling and alignment alone do not fully close the gap.}\\
\textbf{Interpreting Neurons and Features in Models.}
Early work showed that single units in sequence models can represent clear meanings: for example, a single ``sentiment neuron'' appeared in a character-level review generator \citep{seq_neuron}, and manual inspections of character RNNs found neurons that track quotes, brackets, and URLs \citep{recurrent_vis}.
\za{Subsequent} studies identified ``knowledge neurons'' in Transformer-based language models whose activations directly influence cloze completions of factual entities \citep{knowledge_neuron}. 
This complements evidence that MLP layers work like key-value memories that store word and concept associations \citep{ffn_kv}. 
Since many neurons are not tied to just one concept, some work describes their behavior as combinations of multiple features \citep{explain_neuron}. 
At larger scales, some techniques use strong LMs to generate natural-language hypotheses about thousands of neurons and then score them for accuracy \citep{openai2023language}.
For vision-language models (LVLMs), CLIP revealed ``multimodal neurons'' that respond to both text and images for abstract ideas (e.g., celebrities, symbols) \citep{openaimm_neuron}.
Extending ``knowledge neurons'' to LVLMs, a two-stage filtering method found units linked to world knowledge in MiniGPT-4 and enabled targeted edits during image captioning \citep{LMM_knowledge_neuron}. 
Other recent studies also explore safety neurons  that affect alignment-related refusal or harmful outputs \citep{safe_neuron}, and multilingual neurons that hold language-specific competence and can steer which language the model uses \citep{multiling}.
Overall, these studies suggest that neurons provide a useful basis for understanding the abilities of modern models.
\za{However, no prior work has investigated whether inference-related neurons are shared across LLMs and LVLMs, nor how such shared circuitry can be leveraged for cross-model transfer—a gap our work aims to fill.}

\section{Preliminary}

\paragraph{Transformer-based language models.}
A Transformer block consists of a multi-head attention (MHA) module and a feed-forward network (FFN or MLP), both operating on the residual stream.  
Given an input sequence $w = \langle w_0,\ldots,w_t\rangle$, token embeddings $h_i$ are produced by $W_E$ and accumulated in the residual stream.  
Each block then refines this stream (layer normalization omitted):
\begin{equation}
h_{i}^{\ell+1} = h_{i}^{\ell} + \mathrm{MHA}^{\ell}(h_{i}^{\ell}) + \mathrm{MLP}^{\ell}\!\big(h_{i}^{\ell} + \mathrm{MHA}^{\ell}(h_{i}^{\ell})\big).
\end{equation}

\paragraph{Feed-forward neurons.}  
In modern Transformers, the feed-forward network (FFN), or multi-layer perceptron (MLP), can be expressed as  
\begin{equation}
\mathrm{MLP}(x) = W_{\mathrm{down}}^\top \sigma \bigl(W_{\mathrm{up}} x \bigr),
\end{equation}
where $W_{\mathrm{up}}, W_{\mathrm{down}} \in \mathbb{R}^{d_m \times d}$.  
Here $W_{\mathrm{up}}$ projects the residual stream into a higher-dimensional space (\texttt{fwd.up}), producing intermediate activations.  
A non-linear activation $\sigma(\cdot)$ is then applied, and $W_{\mathrm{down}}$ projects the result back to the model dimension $d$ (\texttt{fwd.down}).

\paragraph{Attention neurons.}
The multi-head attention module computes token interactions via four projection matrices:
\begin{align}
Q &= W_Q h, & K &= W_K h, &
V &= W_V h,  
\end{align}
where $W_Q, W_K, W_V \in \mathbb{R}^{d\times d}$ are the query, key, and value, respectively.  
Analogous to the MLP case, each row or column of these matrices may be regarded as a \emph{neuron}:
- \texttt{attn.q}: query neurons that encode how tokens \emph{ask for} information;  
- \texttt{attn.k}: key neurons that encode how tokens \emph{offer} information;  
- \texttt{attn.v}: value neurons that carry the content to be propagated;

\paragraph{Neuron Types Used in Our Work.}
In this work, we analyze these neurons in both LLMs and LVLMs, focusing on their roles in inference.  
This unified view enables us to compare the neurons  
$\{\texttt{attn.q},\texttt{attn.k},\texttt{attn.v},\texttt{fwd.up},\texttt{fwd.down}\}$  
and to investigate whether shared inference circuitry emerges across modalities.

\section{Exploring Shared inference Basis}
\label{sec:share}
\subsection{Identifying Important Neurons in LLMs and LVLMs}
\paragraph{Constructing Neuron Detection Corpus $\boldsymbol{c}$.}
To identify neurons that are important for inference in both LLMs and LVLMs, we define a neuron detection corpus $c$ as a set of $m$ probe inputs $c=\{\mathbf{x}_i\}_{i=1}^m$, where each $\mathbf{x}_i$ concatenates a problem with the model’s own inference output to elicit genuine inference activations. We explicitly study two categories of base models: \emph{text models} and \emph{vision-language (VL) models}. In the text-only pipeline (GSM8K \citep{gsm8k}), a text model $M_0^{\text{text}}$ generates a rationale $r_i$ and prediction $\hat a_i$ for problem $q_i$, and we build $\mathbf{x}_i^{Q+A}=[\texttt{INST}] \Vert q_i \Vert [\texttt{SEP}] \Vert (r_i,\hat a_i)$.The token \texttt{INST} specifies the instruction prefix that standardizes the output format, and \texttt{SEP} is a separator token that marks the boundary between the original problem and the model-generated inference. The operator ``$\Vert$'' represents sequence concatenation.
In the multimodal pipeline (Geo3K \citep{geo3k}), a frozen VL model $M_0^{\text{vl}}$ consumes both the problem text $q_i$ and image $I_i$, where the image is encoded into visual tokens $v_i=\phi(I_i)$, yielding $\mathbf{x}_i^{VL}=[\texttt{INST}] \Vert v_i \Vert q_i \Vert [\texttt{SEP}] \Vert (r_i,\hat a_i)$. This separation allows us to compare inference neurons activated in textual inference versus those engaged in multimodal inference.

\paragraph{Activation Profiling for Identifying inference Neurons.}
We use activation profiling to identify inference neurons, defined as neurons that are consistently activated when processing inputs from inference tasks. Here, a neuron refers to a single row or column within the model’s parameter matrices. Building on prior work in identifying important neurons in neural networks~\citep{lottery,wang2022finding,safe_neuron}, we consider a neuron to be activated if its removal leads to a significant change in the resulting embedding. Formally, given an input sequence $\mathbf{x}$ from context $c$, a neuron $N$ is considered activated if
\begin{equation}
\|\mathrm{Model}(\mathbf{x}) - \mathrm{Model}_{\ominus N}(\mathbf{x})\|_2 \;\ge\; \sigma,
\label{eq:ident-1}
\end{equation}
where $\mathrm{Model}(\mathbf{x})$ denotes the output embedding when processing $\mathbf{x}$, and $\mathrm{Model}_{\ominus N}(\mathbf{x})$ denotes the output when neuron $N$ is deactivated by zeroing the parameters that produce or consume it. The threshold $\sigma$ specifies the minimum magnitude of change required to consider a neuron activated.
Context-related neurons for a specific context $c$ are then \begin{equation}
\begin{aligned}
\mathcal{N}^{c}_{\text{ctx}}
&:= 
\Big\{\, N \in \mathrm{Model}\;\big|\;
\;\|\mathrm{Model}(\mathbf{x}) 
\\
&- \mathrm{Model}_{\ominus N}(\mathbf{x})\|_2 \ge \sigma, \forall\, \mathbf{x}\in c \Big\}. 
\end{aligned}
\label{eq:ident-2}
\end{equation}

We apply Eq. \ref{eq:ident-1}-\ref{eq:ident-2} separately to the text model $M_0^{\text{text}}$ and the VL model $M_0^{\text{vl}}$.

\paragraph{Inference Basis Neurons.}
Our goal is to identify neurons that consistently support inference across both \textbf{text} and \textbf{vision-language (VL)} models.  Let $\mathcal{C}_{\text{text}}$ and $\mathcal{C}_{\text{vl}}$ denote the sets of contexts from the text-only and multimodal pipelines, respectively. Using Eq. \ref{eq:ident-2}, compute context-related sets for each model, $\mathcal{N}^{c}_{\text{ctx}}(M_0^{\text{text}})$ and $\mathcal{N}^{c}_{\text{ctx}}(M_0^{\text{vl}})$, restricted to neurons with one-to-one correspondence in the shared language backbone. We define the \emph{inference Basis (shared) neurons} as
\begin{equation}
\mathcal{N}_{\text{shared}}
\;:=\;
\Big(\!\bigcap_{c\in\mathcal{C}_{\text{text}}} \mathcal{N}^{c}_{\text{ctx}}(M_0^{\text{text}})\!\Big)
\;\cap\;
\Big(\!\bigcap_{c\in\mathcal{C}_{\text{vl}}} \mathcal{N}^{c}_{\text{ctx}}(M_0^{\text{vl}})\!\Big).
\label{eq:ident-3}
\end{equation}
This definition explicitly measures neurons that support inference across both model classes, while separating those specialized to text-only or multimodal settings.

\paragraph{Neuron-Concept Validation via Amplification.}
To more clearly assess the functional role of each neuron,
we amplify its activation during the forward pass:

\begin{equation}
N \;\leftarrow\; \lambda \cdot N , N \in \mathcal{N}_{\text{shared}}, 
\end{equation}
where $\lambda$  denotes the amplification factor.
For each token $t$, we directly measure its frequency under amplified runs:
\begin{equation}
F_{\lambda}(t) \;=\; \sum_{\mathbf{x}\in c}\sum_j \mathbb{1}\!\left[y^{\oplus \lambda N}_j = t\;\middle|\; x^Q ,y_{<j}\right],    
\end{equation}
where $y^{\oplus \lambda N}_j$ denotes the token generated at position $j$ when neuron $N$ is amplified.

\subsection{Findings of Shared inference Basis}
In this section, we report three findings that together illuminate the shared inference mechanisms underlying diverse model families.
\begin{finding}[RoyalBlue]{}
Different series of VL models and their corresponding text-math models share a large number of neurons on inference tasks.
\end{finding}

\begin{figure*}[t]
    \centering
    \begin{minipage}[t]{0.25\linewidth}
    \centering
    \includegraphics[width=\linewidth]{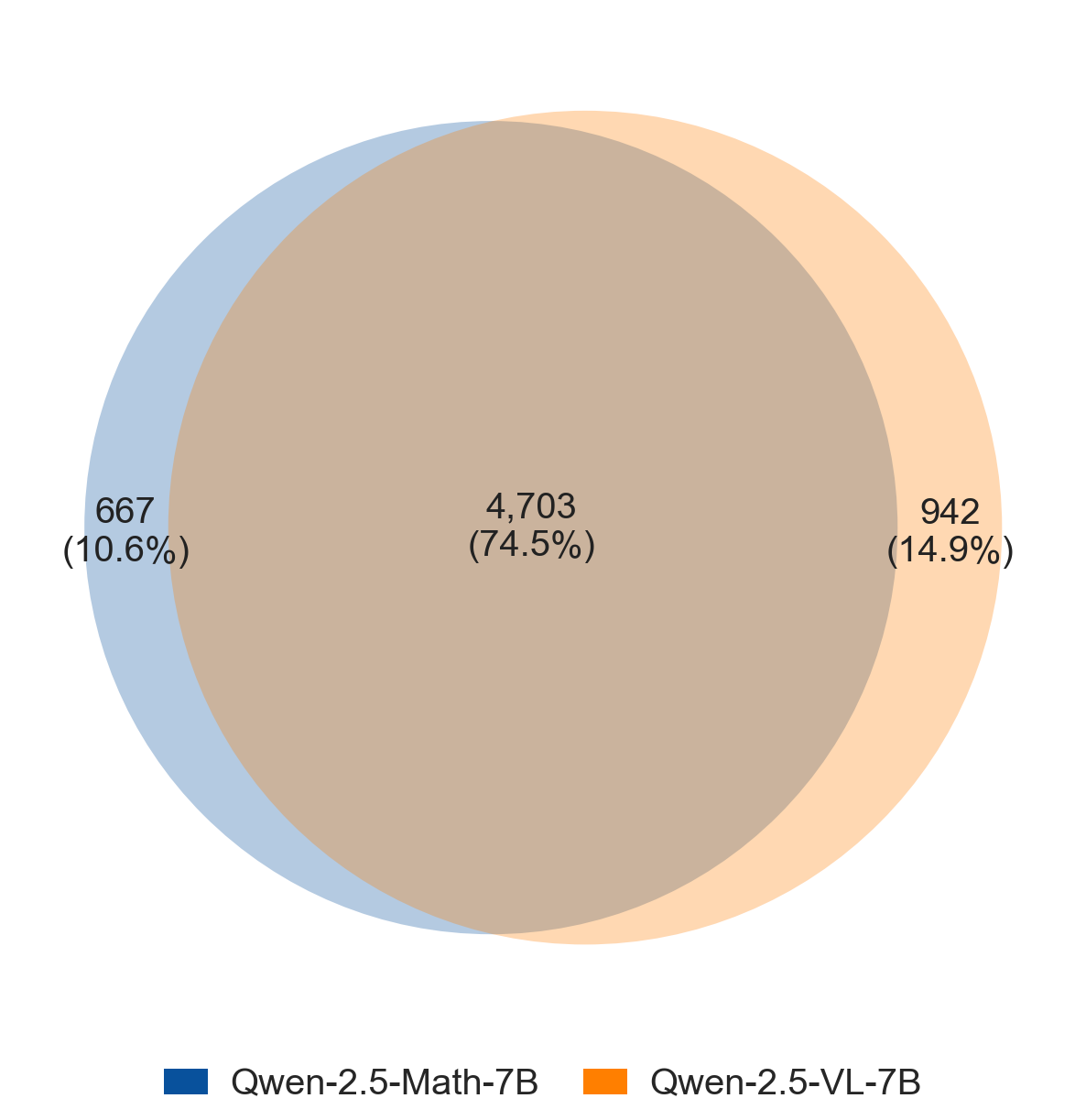}
    \subcaption{Qwen2.5-VL-7B vs Qwen2.5-Math-7B}\label{fig:venn-qwenvl}
  \end{minipage}
      \begin{minipage}[t]{0.25\linewidth}
    \centering
    \includegraphics[width=\linewidth]{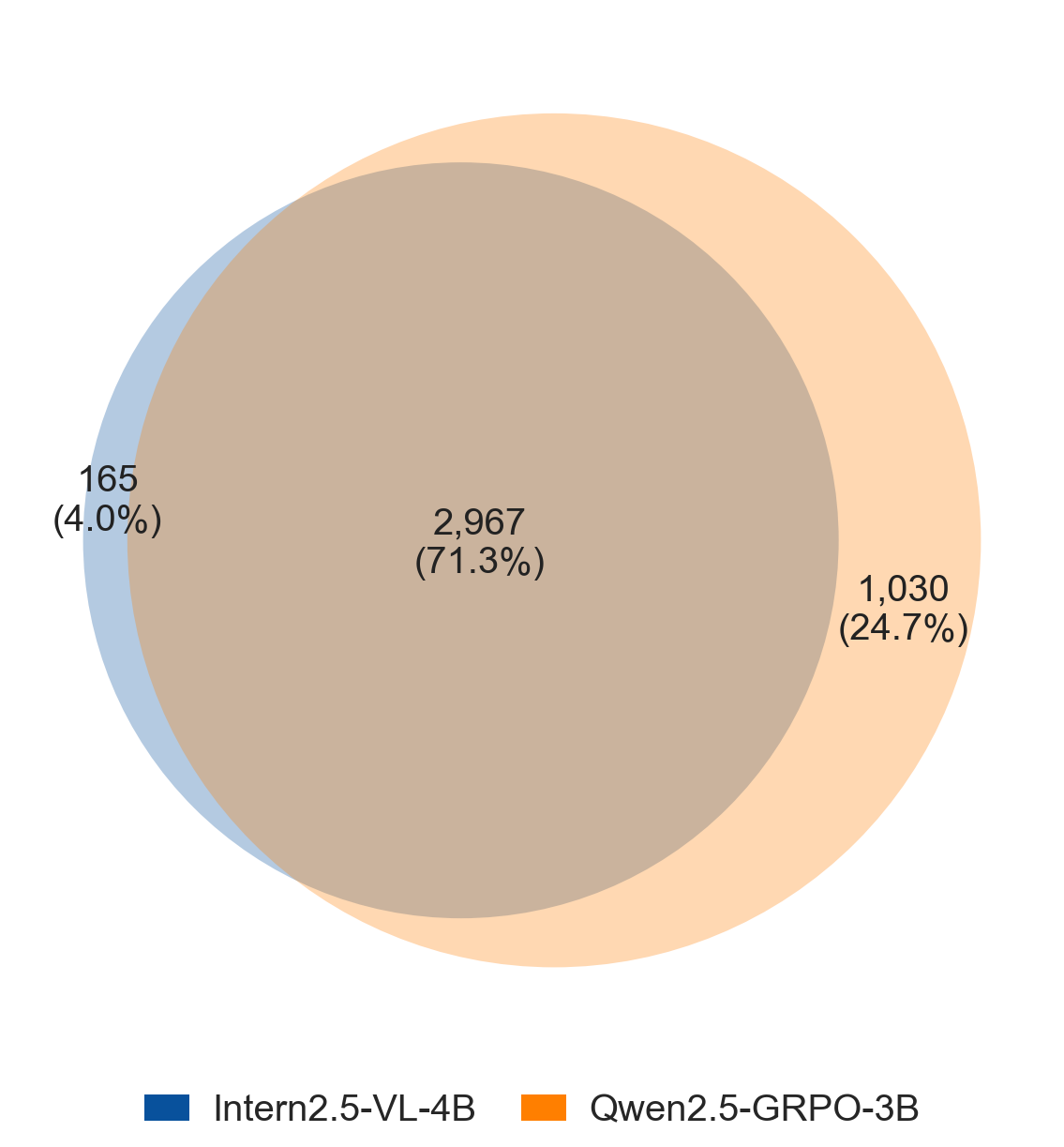}
    \subcaption{Intern2.5-VL-4B vs Qwen2.5-GRPO-3B}\label{fig:venn-intern}
  \end{minipage}
    \begin{minipage}[t]{0.25\linewidth}
    \centering
    \includegraphics[width=\linewidth]{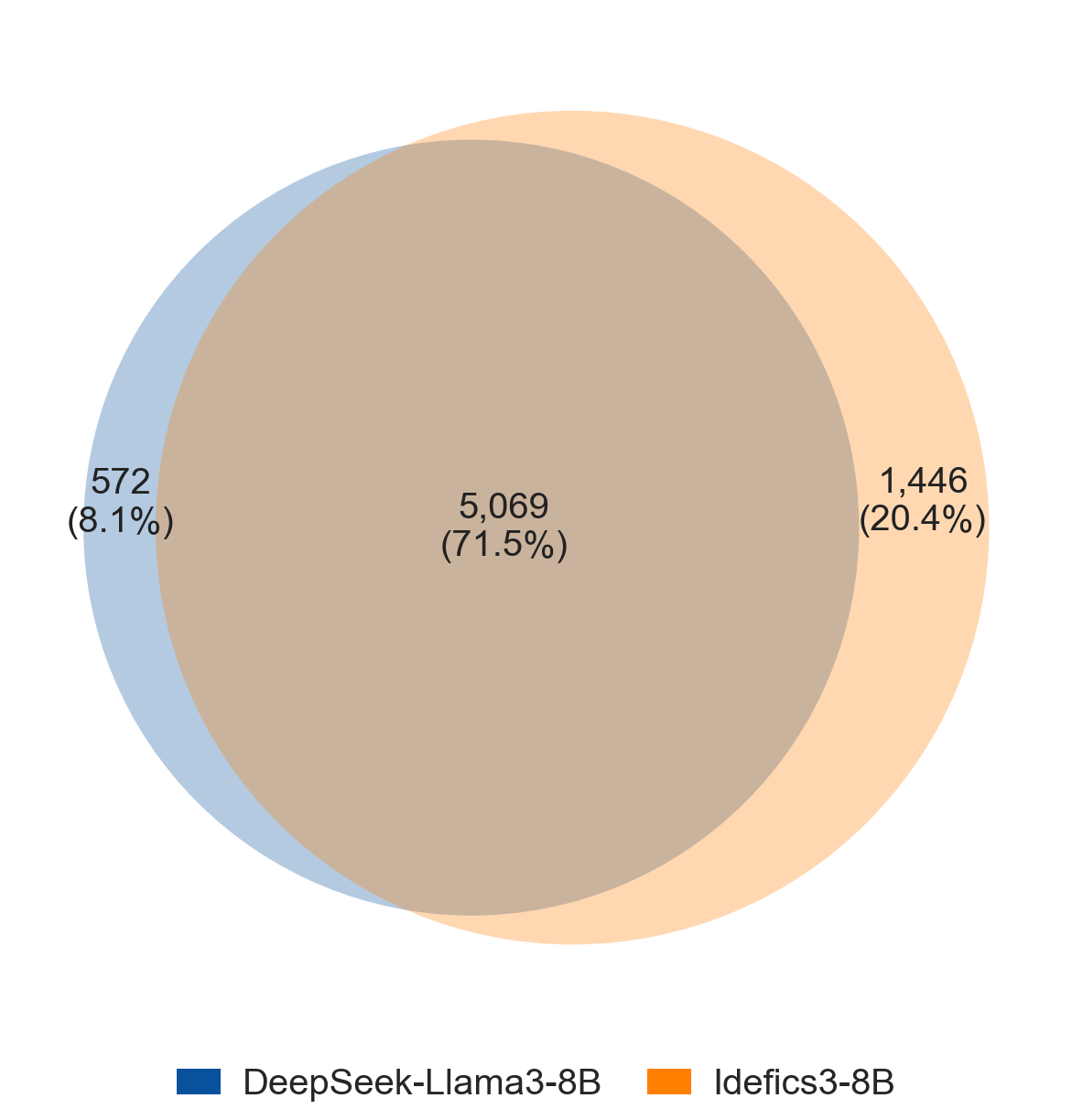}
    \subcaption{Idefics3 vs Deepseek-LLaMA3}\label{fig:venn-idefics}
  \end{minipage}
  \caption{Overlap of activated neurons across different LVLM families.}
  \label{fig:venn_main}
\end{figure*}

\paragraph{Strong overlap.}
Across multiple LLM/LVLM families (e.g., Qwen2.5-VL-7B \citep{qwen2_5vl} vs.\ Qwen2.5-Math-7B \citep{qwen2_5_math};  Intern2.5-VL-3B \citep{internvl2} vs.\ Qwen2.5-GRPO-3B \citep{qwen3b_grpo}; LLaVA-Next-8B \citep{llavanext} vs.\ LLaMA3-DeepSeek-8b \citep{deepseekr1}), we observe a substantial intersection between the sets of neurons activated by inference inputs.
As shown in Figure~\ref{fig:venn_main}, the Qwen2.5-7B-Math and Qwen2.5-7B-VL share \emph{4{,}703} neurons out of a union of \emph{6{,}312} (\textbf{74.5\%} of the union), with only \emph{667} (10.6\%) and \emph{942} (14.9\%) neurons remaining model-specific.  Similar levels of overlap are also observed in the Idefics3-8B.
Interestingly, we further find that even models without identical language backbones still exhibit a non-negligible proportion of shared neurons, suggesting the existence of universal inference units that transcend architectural differences. See details in Appendix \ref{app:overlap}.

This magnitude of overlap is consistently seen across other model pairs we tested (details in Appendix), indicating that a nontrivial portion of the inference circuitry is reused between text-only and multimodal models despite their different training signals.

\begin{figure}[h]
    \centering
    \includegraphics[width=\linewidth]{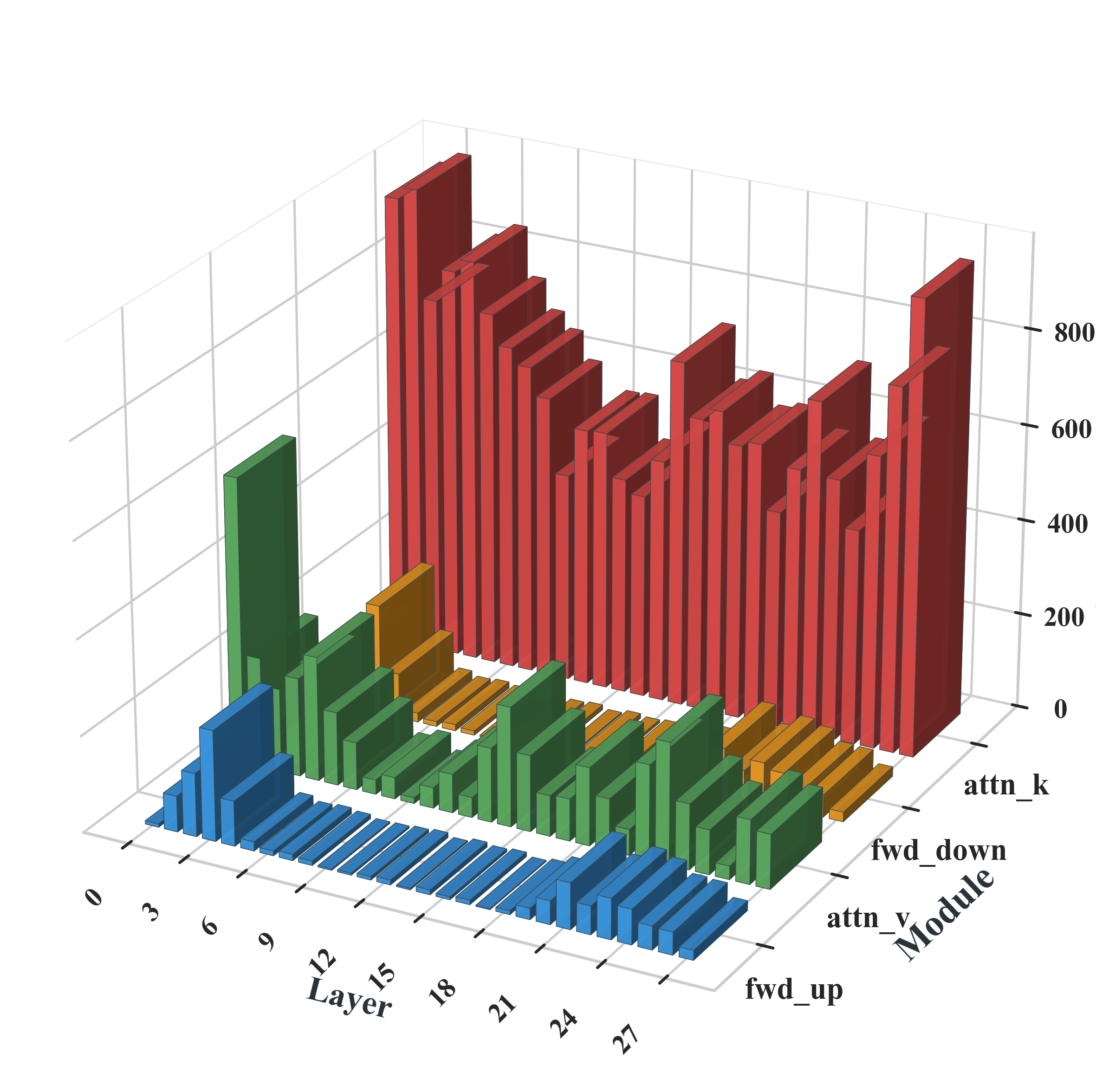}
    \caption{Distribution of shared neurons across layers and modules in Qwen2.5-VL-7B and Qwen2.5-Math-7B. 
    A clear concentration appears in \texttt{attn.k}, with secondary presence in \texttt{attn.v}. 
    Clustering is most evident in early layers  and later layers.}
    \label{fig:shared-layer-module}
\end{figure}

\textbf{Where the shared neurons live.}
We next examine how the shared neurons distribute across layers and modules.
As shown in Figure~\ref{fig:shared-layer-module}, the shared neurons in Qwen2.5-VL-7B and Qwen2.5-Math-7B are clearly concentrated in the {\texttt{attn.k}} matrices across nearly all layers, with a secondary presence in \texttt{attn.v}.
Additionally, distinct clustering emerges in the early layers (e.g., before layer 6), and again in later layers (after layer 20). Further examples are provided in Appendix~\ref{app:overlap}.

\begin{finding}[RoyalBlue]{}
Shared inference basis plays an important role in a model’s performance on inference tasks.
\end{finding}

We evaluate the \emph{causal} role of shared neurons ($\mathcal{N}_{\text{shared}}$; Eq.~\ref{eq:ident-3}) by (i) \textbf{Deact}—zeroing the parameters that produce/consume only neurons in $\mathcal{N}_{\text{shared}}$; and (ii) \textbf{Random Deact}—ablating the \emph{same number} of neurons sampled at random from the same layer-module budget.

\begin{figure}
    \centering
    \includegraphics[width=\linewidth]{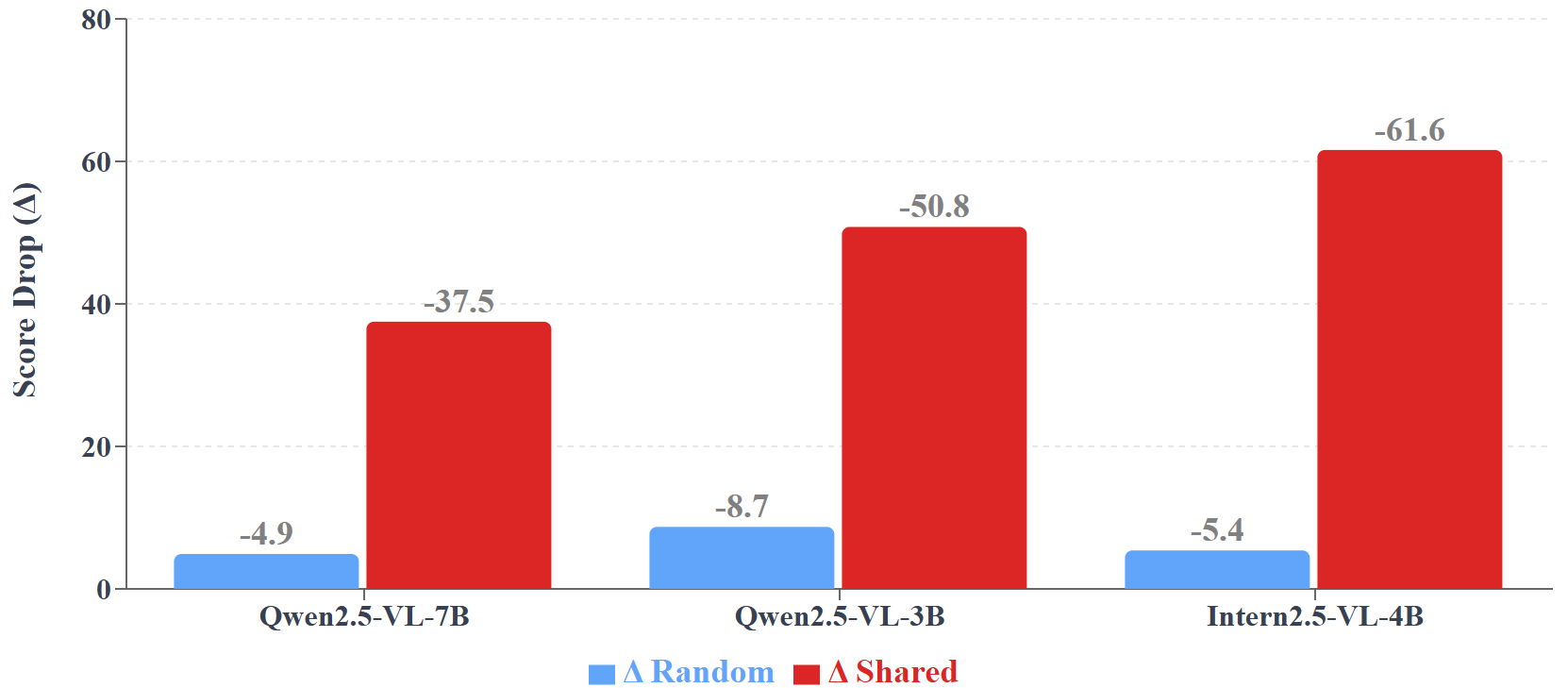}
\caption{\textbf{Effect of ablating shared neurons on \textsc{MathVista}.  
Arrows indicate higher is better.}}
    \label{fig:mathvista_testmini_solution}
\end{figure}

Table~\ref{fig:mathvista_testmini_solution} summarizes the effect on accuracy in MathVista~\citep{lu2023mathvista}, 
a comprehensive benchmark for mathematical inference in multimodal large language models. 
Across three LVLMs, deactivating the shared neurons collapses performance to \textbf{0.0}, whereas random ablations of equal size lead to notably smaller drops. Parallel trends appear across additional benchmarks (see Appendix~\ref{app:finding2}), indicating that the shared neurons are \emph{both necessary and specific} to inference. The dramatic failure under \textbf{Deact}—in sharp contrast to the partial degradation under \textbf{Random Deact}—shows that the \emph{shared inference basis} encodes functionally indispensable circuitry for multimodal inference rather than generic capacity or redundant pathways. Further results on a wider suite of tasks (Appendix~\ref{app:finding2}) reinforce this pattern, underscoring the centrality of these shared neurons to robust inference behavior.
 
\begin{figure}[htbp]
    \centering
    \begin{subfigure}[b]{0.31\linewidth}
        \centering
        \includegraphics[width=\linewidth]{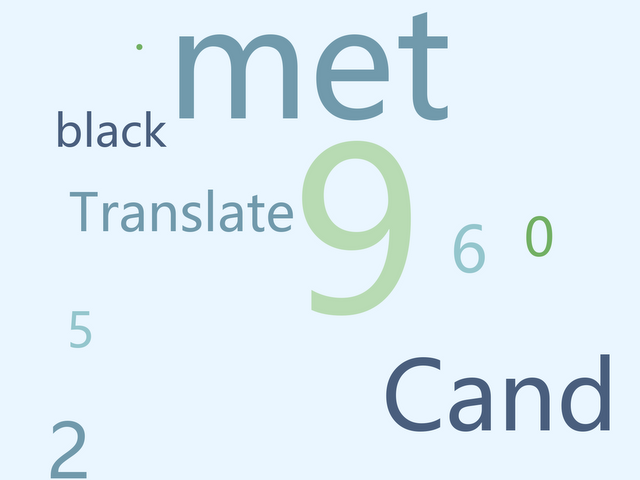}
        \caption*{\scriptsize \texttt{fwd.up.L14.N12953}}
    \end{subfigure}
    \hfill
    \begin{subfigure}[b]{0.31\linewidth}
        \centering
        \includegraphics[width=\linewidth]{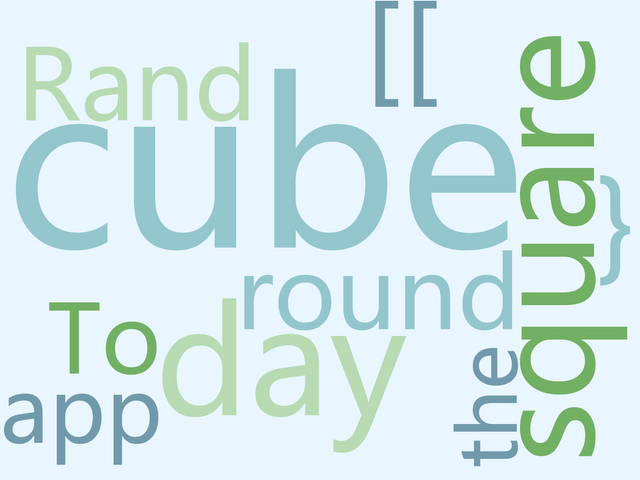}
        \caption*{\scriptsize \texttt{fwd.down.L10.N1889}}
    \end{subfigure}
    \hfill
    \begin{subfigure}[b]{0.31\linewidth}
        \centering
        \includegraphics[width=\linewidth]{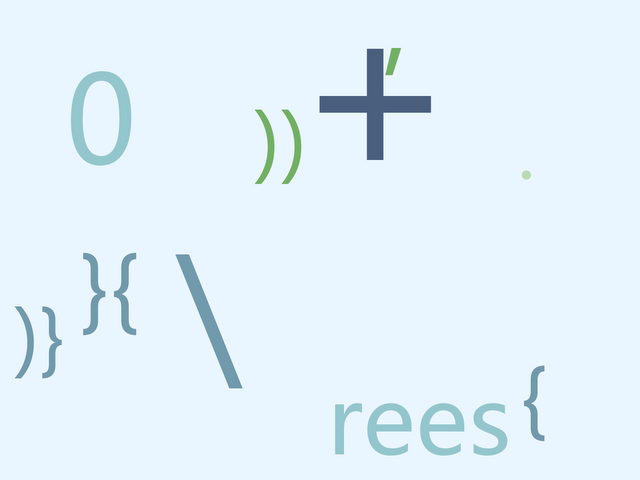}
        \caption*{\scriptsize \texttt{attn.v.L1.N2093}}
    \end{subfigure}

    \caption{Word clouds of token concepts for different neurons.}
    \label{fig:wordclouds_main}
\end{figure}

\begin{finding}[RoyalBlue]{}
A large portion of shared neurons are linked to math inference.
\end{finding}
Using our amplification method (§3), we find shared neurons in both LLMs and LVLMs that show clear activation on mathematical concepts. To better illustrate this, we build word clouds from the amplified generations of Qwen2.5-Math-7B, where token substrings are mapped back to their original text. Word size shows activation frequency, making the main math tendencies of each neuron visible. As shown in Figure~\ref{fig:wordclouds_main}, different neurons are sensitive to different parts of math inference. For example, neuron \texttt{fwd.up.L14.N12953} activates on algebra and arithmetic tokens (e.g., numbers, ``met,''), while neuron \texttt{fwd.down.L10.N1889} responds to geometry and measurement concepts (e.g., ``cube,'' ``square,'' ``round''). These patterns suggest that some neurons specialize in certain types of mathematical knowledge. This finding shows an important property of large models: internal neurons can act as carriers of math inference, helping explain how inference ability emerges.
 Additional examples are provided in Appendix \ref{app:finding3}.

 

 \section{Merging Technique}
\label{sec:method}

We propose a parameter-efficient \emph{Shared Neuron Low-Rank Fusion (SNRF)} that transfers inference circuitry between two models by \textbf{selectively} updating only neurons empirically identified as \emph{shared} (Section \ref{sec:share}). In our method there are exactly two models: a \emph{source}  text model and a  \emph{target} vision-language model (VL). For every transformer layer $\ell$ and projection type $p\in\{\texttt{attn.q},\texttt{attn.k},\texttt{attn.v},\texttt{fwd.up},\texttt{fwd.down}\}$, let
$W^{\mathrm{src}}_{\ell,p},W^{\mathrm{tgt}}_{\ell,p}\in\mathbb{R}^{m_{\ell,p}\times n_{\ell,p}}$
denote the corresponding weight matrices. Let $\mathcal{S}^{\ell,p}$ be the index set of \emph{shared neurons} for $(\ell,p)$, which forms the basis on which SNRF applies its low-rank fusion update.

\subsection{Problem Setup}
Our goal is to construct merged weights $\widetilde{W}_{\ell,p}$ that  
(i) preserve the target model parameters outside $\mathcal{S}^{\ell,p}$, and  
(ii) inject a compact, low-rank transfer signal only on the shared-neuron subset $\mathcal{S}^{\ell,p}$.  

We begin by defining the inter-model difference:
\begin{equation}
\Delta_{\ell,p} := W^{\mathrm{src}}_{\ell,p} - W^{\mathrm{tgt}}_{\ell,p}.
\label{eq:delta}
\end{equation}
Let $M_{\mathcal{S}^{\ell,p}}(\cdot)$ denote a masking operator that keeps only the entries indexed by $\mathcal{S}^{\ell,p}$ and sets all others to zero.  

\subsection{Selective Low-Rank Update}
We obtain a rank-$r$ approximation of $\Delta_{\ell,p}$ using SVD:
\begin{equation}
\begin{aligned}
\Delta_{\ell,p} &= U_{\ell,p}\Sigma_{\ell,p}V_{\ell,p}^{\top}, \\[3pt]
\Delta^{(r)}_{\ell,p} &:= U_{\ell,p}[:,1{:}r]\;\Sigma_{\ell,p}[1{:}r,1{:}r]\;V_{\ell,p}[:,1{:}r]^{\top},
\end{aligned}
\label{eq:tsvd}
\end{equation}

where $r \ll \min(m_{\ell,p}, n_{\ell,p})$.  

The merged weight is then
\begin{equation}
\widetilde{W}_{\ell,p} =
W^{\mathrm{tgt}}_{\ell,p}
+ \beta\, M_{\mathcal{S}^{\ell,p}}\!\bigl(\Delta^{(r)}_{\ell,p}\bigr),
\label{eq:update}
\end{equation}
where $\beta \in [0,1]$ controls the strength of the update.  

\subsection{Comparison with Linear Parameter Merging}
\label{sec:theory}

We compare \textsc{SNRF} with linear parameter merging and show that,
for small $\beta$, masking the update outside the shared-inference
subspace $S$ yields smaller loss whenever curvature in
$S^{\perp}$ dominates the in-$S$ truncation error.

\paragraph{Main Result.}
Let $\Delta = W^{\src}-W^{\tgt}$ and decompose
$\Delta_S = P_S\Delta$, $\Delta_{\perp}=P_{\perp}\Delta$.
\textsc{SNRF} applies a masked low-rank update
$\Delta^{(r)}_S$, while linear merging applies the full update.
For sufficiently small $\beta$,
\begin{equation}
\label{eq:gap}
\begin{aligned}
\Delta\mathcal{L}&_{\lin}(\beta)-\Delta\mathcal{L}_{\snrf}(\beta)
\;\ge\;
\frac{\beta^{2}}{2}\mu_{\perp}\|\Delta_{\perp}\|_{F}^{2}-
\\[-2pt]
&\!\!\!\!\!\!\!\!\!\! 
\frac{\beta^{2}}{2}\mu_{S}\|\Delta_{S}-\Delta_{S}^{(r)}\|_{F}^{2}
-
\beta\,c\,\|P_{S}g\|_{F}\,\|\Delta\|_{F}
+O(\beta^{3}) .
\end{aligned}
\end{equation}

where $c=\varepsilon(1+\eta)$ is small.
Thus, linear merging incurs an unavoidable penalty from curvature in
$S^{\perp}$, while \textsc{SNRF} masks this component and pays only the
rank-$r$ truncation cost inside~$S$.

Whenever
\begin{equation}
\mu_{\perp}\|\Delta_{\perp}\|_F^2
\;>\;
\mu_S\|\Delta_S-\Delta^{(r)}_S\|_F^2
\;+\;
c\,\beta^{-1} \|P_S g\|_F\,\|\Delta\|_F,
\label{eq:cond}
\end{equation}
we obtain the strict improvement
\[
\Delta\mathcal{L}_{\snrf}(\beta)
<
\Delta\mathcal{L}_{\lin}(\beta).
\]

Detailed derivations are provided in Appendix~\ref{app:proof}.

\begin{algorithm}[t]
\caption{Shared Neuron Low-Rank Fusion (SNRF)}
\label{alg:snrf}
\begin{algorithmic}[1]
\Require Source weights $\{W^{\mathrm{src}}_{\ell,p}\}$; 
        Target weights $\{W^{\mathrm{tgt}}_{\ell,p}\}$; 
        shared index sets $\{\mathcal{S}^{\ell,p}\}$; \\
        \hspace{1.5em} rank $r$; mixing coefficient $\beta$
\For{each layer $\ell$ and projection $p$}
  \If{$\mathcal{S}^{\ell,p}\neq\varnothing$}
    \State $\Delta_{\ell,p}\gets W^{\mathrm{src}}_{\ell,p}-W^{\mathrm{tgt}}_{\ell,p}$ \Comment{inter-model delta}
    \State $(U,\Sigma,V)\gets \textsc{SVD}(\Delta_{\ell,p})$
    \State $\Delta^{(r)}_{\ell,p}\gets U[:,1{:}r]\;\Sigma[1{:}r,1{:}r]\;V[:,1{:}r]^{\top}$ \Comment{rank-$r$ approx.}
    \State $\widetilde{W}_{\ell,p}\gets W^{\mathrm{tgt}}_{\ell,p} + \beta\, M_{\mathcal{S}^{\ell,p}}\!\big(\Delta^{(r)}_{\ell,p}\big)$ \Comment{selective update}
  \Else
    \State $\widetilde{W}_{\ell,p}\gets W^{\mathrm{tgt}}_{\ell,p}$
  \EndIf
\EndFor
\State \Return merged weights $\{\widetilde{W}_{\ell,p}\}$
\end{algorithmic}
\end{algorithm}

\section{Experiments}
\label{sec:experiments}
We conduct experiments to answer the following research questions:  
(1) How does our method perform compared to the original model?  
(2) How does our method perform compared to other baselines? 
(3) How well does our method generalize to other parameter-merging techniques?
\begin{table*}[!t]
\centering
\caption{\textbf{Main results of our Shared Neuron Low-Rank Fusion (SNRF) across backbones.}
Numbers in parentheses for \textit{Ours} denote absolute deltas vs.\ the corresponding baseline. Higher is better for all metrics shown.}
\renewcommand{\arraystretch}{1.12}
\resizebox{\textwidth}{!}{%
\begin{tabular}{lcccccccc}
\toprule
\rowcolor{lightgray}
\textbf{Model} &
\textbf{CoT $\uparrow$} &
\textbf{Format $\uparrow$} &
\textbf{Solution $\uparrow$} &
\textbf{MME $\uparrow$} &
\textbf{POPE $\uparrow$} &
\textbf{ScienceQA $\uparrow$} &
\textbf{MMMU (val $\uparrow$)} &
\textbf{MMMU-Pro (V $\uparrow$)}\\
\midrule

Qwen2.5-VL-3B & 52.6 & 61.6 & 50.8 & 1535.0 & 87.2 & 52.5 & 0.461 & 0.018 \\
\rowcolor{lightgray}
\textit{Ours} & 55.2 \,(+2.6) & 57.9 \,(-3.7) & 55.5 \,(+4.7) & 1559.0 \,(+24.0) & 88.0 \,(+0.8) & 75.1 \,(+22.6) & 0.458 \,(-0.003) & 0.070 \,(+0.052) \\
\midrule

Qwen2.5-VL-7B & 48.8 & 68.8 & 37.5 & 1681.0 & 86.2 & 54.5 & 0.503 & 0.116 \\
\rowcolor{lightgray}
\textit{Ours} & 53.3 \,(+4.5) & 68.1 \,(-0.7) & 42.1 \,(+4.6) & 1713.0 \,(+32.0) & 86.5 \,(+0.3) & 55.4 \,(+0.9) & 0.503 \,(-0.000) & 0.179 \,(+0.063) \\
\midrule

Intern2.5-VL-4B & 60.4 & 65.1 & 61.6 & 1670.0 & 90.8 & 97.4 & 0.491 & 0.000 \\
\rowcolor{lightgray}
\textit{Ours} & 60.6 \,(+0.2) & 65.9 \,(+0.8) & 61.3 \,(-0.3) & 1677.0 \,(+7.0) & 90.7 \,(-0.1) & 97.3 \,(-0.1) & 0.486 \,(-0.005) & 0.186 \,(+0.186) \\
\midrule

Idefics3-8B-LLaMA3 & 50.1 & 50.4 & 50.3 & 1458.0 & 86.0 & 25.9 & 0.423 & 0.084 \\
\rowcolor{lightgray}
\textit{Ours} & 50.4 \,(+0.3) & 49.6 \,(-0.8) & 51.1 \,(+0.8) & 1460.0 \,(+2.0) & 84.9 \,(-1.1) & 40.5 \,(+14.6) & 0.426 \,(+0.002) & 0.097 \,(+0.013) \\
\midrule

LLaVA-Next-8B & 35.8 & 38.6 & 36.0 & 1586.0 & 87.2 & 74.4 & 0.410 & 0.051 \\
\rowcolor{lightgray}
\textit{Ours} & 36.5 \,(+0.7) & 38.1 \,(-0.5) & 36.2 \,(+0.2) & 1622.0 \,(+36.0) & 87.5 \,(+0.3) & 74.6 \,(+0.2) & 0.400 \,(-0.010) & 0.051 \,(+0.000) \\
\bottomrule
\end{tabular}}
\label{tab:main-results-base}
\end{table*}

\subsection{Setup}
\paragraph{Backbones.}
Unless otherwise noted, we evaluate LVLMs at small/medium scales (3B--8B) built on two language backbones: 
(i) \textbf{Llama~3}---\emph{Idefics3-8B-LLaMA3} \citep{idefics3} and \emph{LLaVA-Next-8B} \citep{llavanext}, where the corresponding text model used for merging is \emph{DeepSeek-LLaMA3} \citep{deepseekr1}; 
(ii) \textbf{Qwen2.5}—We evaluate \emph{Qwen2.5-VL-3B},   \emph{Qwen2.5-VL-7B} \citep{qwen2_5vl}, and \emph{Intern2.5-VL-4B} \citep{internvl2}. For parameter merging, \emph{Qwen2.5-VL-7B}  is paired with \emph{Qwen2.5-Math} \citep{qwen2_5_math}, whereas  \emph{Intern2.5-VL-4B} and \emph{Qwen2.5-VL-3B} are paired with the text backbone \emph{Qwen3B-GRPO} \citep{qwen3b_grpo}. Each model uses its official vision encoder as released by the authors. Exact model and checkpoints are listed in Appendix \ref{app:models}.

\paragraph{Benchmarks.}
We include MathVista \citep{lu2023mathvista} for visual mathematical inference, \emph{MME} \citep{mme} for broad perception and cognition, \emph{POPE} \citep{pope} for hallucination detection, and \emph{ScienceQA} \citep{sciqa} for multimodal scientific problem solving with explanations. Together, these benchmarks allow us to measure modality alignment, perception$\to$inference competence, and safety/calibration under diverse conditions. We evaluate models on both the original \emph{MMMU} \citep{mmmu} and the more challenging \emph{MMMU-Pro} \citep{mmmu_pro}.  
\emph{MMMU} evaluates college-level, multi-discipline multimodal understanding, while \emph{MMMU-Pro} introduces harder questions and stricter protocols for more robust cross-domain inference. Full dataset splits, preprocessing details, and license information are provided in Appendix \ref{app:benchmarks}.

\paragraph{Baselines.}
We compare our approach against a diverse set of publicly available LVLM baselines at both the 3B and 7B scales.
These baselines all involve training strategies designed to enhance inference ability, including supervised fine-tuning and reinforcement learning, including: 
Qwen2.5-VL CoT-SFT (Training via chain of thought supervised finetuning) \citep{cot-sft},
Qwen2.5-VL Open-R1-Distill  \citep{trl_r1_distill},
VLAA-thinker \citep{r1_rlvlm},
OMlab-VL-Math \citep{vlm_r1_omlab},
Qwen2.5-VL GRIT \citep{GRIT}.
For our method, we additionally explore merging-based training, which is reported separately.
We also experiment with other merging approaches such as linear merge~\citep{model_soup}, DARE~\citep{Dare}, and FRANK~\citep{Frank} to merge shared neurons of LLMs and LVLMs.

\begin{figure*}
    \centering
    \includegraphics[width=0.75\linewidth]{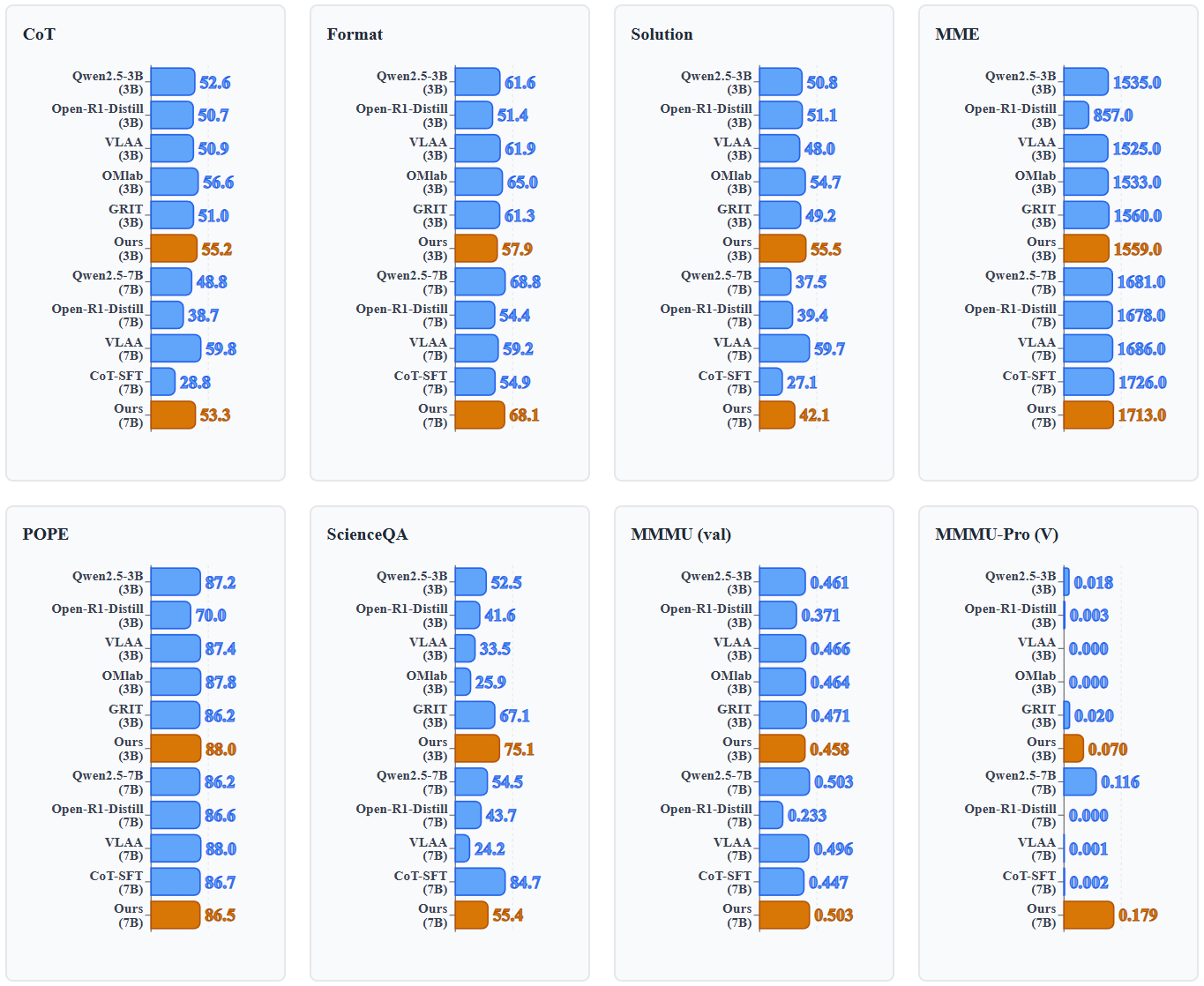}
    \caption{Results on MathVista-TestMini (CoT/Format/Solution), MME, POPE, ScienceQA, MMMU (val) and MMMU-Pro (V) for 3B and 7B scales. Note that Qwen2.5-VL-7B-CoT-SFT, OMlab-VL-3B-Math and Qwen2.5-VL-3B-GRIT do not have corresponding 3B and 7B versions, so their results are not reported.}
    \label{fig:main-results-train}
\end{figure*}

\begin{table*}[t]
\caption{Generalizability of merge strategies.  
Merging shared neurons—whether by our method or alternative schemes—consistently strengthens verification-style inference (\emph{MMMU-Pro (V)}) while preserving perception and hallucination performance. Our \textit{Ours} achieves the most pronounced verification gains.}
\centering
\resizebox{\textwidth}{!}{%
\begin{tabular}{lcccccccc}
\toprule
\rowcolor{lightgray}
\textbf{Baseline} & \textbf{CoT} & \textbf{Format} & \textbf{Solution} & \textbf{MME} & \textbf{POPE} & \textbf{ScienceQA} & \textbf{MMMU (val)} & \textbf{MMMU-Pro (V)} \\
\midrule

\rowcolor{lightgray}
Qwen2.5-VL-7B & 48.8 & \textbf{68.8} & 37.5 & 1681.0 & 86.2 & 54.5 & 0.503 & 0.116 \\
\quad +Linear & 52.4 & 61.3 & \textbf{44.1} & 1669.0 & {86.4} & 56.4 & 0.450 & 0.165 \\
\quad +DARE & 49.7 & 65.6 & 43.2 & 1677.0 & 86.3 & \textbf{58.2} & \textbf{0.511} & 0.125 \\
\quad +FRANK & 49.4 & 68.3 & 42.7 & 1667.0 & \textbf{86.6} & 57.2 & 0.493 & 0.107 \\
\midrule

\rowcolor{lightgray}
Intern2.5-VL-4B & 60.4 & 65.1 & 61.6 & 1670.0 & 90.8 & \textbf{97.4} & 0.491 & 0.000 \\
\quad +Linear & 60.2 & 65.8 & 61.8 & 1670.0 & 90.8 & 97.2 & \textbf{0.493} & \textbf{0.209} \\
\quad +DARE & 59.7 & 65.6 & \textbf{63.2} & 1669.0 & \textbf{90.9} & \textbf{97.4} & 0.486 & 0.181 \\
\quad +FRANK & 59.2 & 65.7 & 61.1 & 1667.0 & 86.6 & 97.3 & \textbf{0.493} & 0.196 \\
\bottomrule
\end{tabular}
}
\label{tab:merging-baselines}
\end{table*}

\subsection{Results and Analysis.}

\paragraph{Comparison with original models.}
As summarized in Table~\ref{tab:main-results-base}, our selective fusion on shared neurons yields consistent improvements over the original models on \emph{inference-centric} benchmarks while largely preserving perception and hallucination behavior. 
On \textbf{MathVista}, we observe an average gain of \textbf{+2.0} points on \emph{Solution} and \textbf{+1.66} on \emph{CoT}, with the largest boosts on Qwen2.5-VL-3B (+4.7 \emph{Solution}, +2.6 \emph{CoT}) and Qwen2.5-VL-7B (+4.6 \emph{Solution}, +4.5 \emph{CoT}). The \emph{Format} score slightly decreases on average (-0.98), indicating that our method primarily transfers \emph{inference} rather than output formatting; this can be mitigated with a lightweight formatting adapter.
On \textbf{MMMU/Pro}, the stricter \emph{MMMU-Pro (V)} improves across backbones by \textbf{+0.0627} on average, with a notable jump on Intern2.5-VL-4B (0.000 $\rightarrow$ \textbf{0.186}). In contrast, \emph{MMMU (val)} remains essentially flat, suggesting that our gains concentrate on harder, verification-style settings rather than inflating validation averages.
For \textbf{ScienceQA}, where textual inference is prominent, we see sizable improvements (\textbf{+7.64} on average): Qwen2.5-VL-3B rises by +22.6 and Idefics3-8B-LLaMA3 by +14.6, supporting our hypothesis that shared neurons encode reusable math/science primitives transferable from text to multimodal contexts.
Perception and hallucination metrics remain stable. The aggregate \textbf{MME} score increases by \textbf{+20.2} on average (e.g., +36 on LLaVA-Next-8B and +32 on Qwen2.5-VL-7B), while \textbf{POPE} is effectively unchanged. This stability is consistent with our design: only neurons empirically identified as \emph{shared inference} units are updated.

 \paragraph{Comparison with training baselines.}
As shown in Figure \ref{fig:main-results-train}, \emph{Ours (Merge)} is strongest on  {inference-intensive} metrics while preserving perception. At 3B, it leads on \textbf{MathVista-Solution} (\textbf{55.5}) and \textbf{ScienceQA} (\textbf{75.1}), and stays near SOTA on \textbf{MME}/\textbf{POPE} (1559/88.0 vs. 1560/88.3). At 7B, it achieves the best \textbf{MMMU-Pro (V)} \textbf{0.179} (prev. best 0.1162) and essentially ties \textbf{MMMU (val)} (0.503 vs. 0.5033). Compared with the corresponding Qwen2.5-VL baselines, \textbf{CoT/Solution} improve at both scales (3B: +2.6/+4.7; 7B: +4.5/+4.6). Remaining gaps are mainly on \textbf{formatting-heavy} metrics (MathVista-Format) and 7B \textbf{ScienceQA} versus SFT-specialized models, while perception and hallucination metrics remain stable.


\paragraph{Generalizability of merging strategies.}
Table \ref{tab:merging-baselines} evaluates the generalizability of our shared-neuron merge strategy. 
On  {Qwen2.5-VL-7B}, applying \emph{Merge} yields \emph{MMMU-Pro (V)} 0.179 (vs.\ Linear 0.165, DARE 0.125, FRANK 0.107) and MME 1713, while keeping \emph{MMMU (val)} stable (0.503) and POPE stable (86.5); MathVista reaches 53.3/68.1/42.1 (CoT/Format/Solution).  
On  {Intern2.5-VL-4B}, \emph{Merge} achieves MathVista 60.6/65.9/61.3 and \emph{MMMU-Pro (V)} 0.186, with \emph{MMMU (val)} (0.486) and perception metrics (MME 1677, POPE 90.7) maintained.  
These results show that merging shared neurons is a generally effective strategy: different merge formulations such as Linear, DARE, and FRANK can also improve abilities of LVLMs.


\section{Conclusion}
This work uncovers a shared neuron subspace between LLMs and LVLMs, showing that many of the most active neurons overlap across modalities and encode causal, concept-level semantics. Leveraging this insight, we introduce \textbf{Shared Neuron Low-Rank Fusion (SNRF)}, a parameter-efficient method that transfers mature textual abilities into LVLMs by selectively aligning and updating these shared neurons. SNRF consistently improves multimodal inference while preserving perception and hallucination performance, outperforming training-based and merging baselines. These findings suggest that LLMs and LVLMs share internal inference circuitry, offering a pathway for cross-domain knowledge transfer—such as from code or math specialists to multimodal models.


{
    \small
    \bibliographystyle{ieeenat_fullname}
    \bibliography{main}
}



\appendix 
\section{Neuron Detection Algorithm}

Inspired by prior work \citep{safe_neuron,multiling}, we adopt a \emph{parallelizable neuron detection method}. 
Unlike the main definition in Eq.~(4), which considers the change in the \emph{final} output embedding after deactivating a neuron, here we compute the impact \emph{locally at the layer containing the neuron}. 
This layer-wise impact serves as a proxy (or component) for the overall impact while being far more efficient to compute.

Formally, let $X \in \mathbb{R}^{l \times d_{\text{model}}}$ denote the hidden states input to a given layer, 
where $l$ is the sequence length and $d_{\text{model}}$ the hidden dimension. 
For a neuron $N$ in this layer, its impact is measured as:
\begin{equation}
\| f(X;\theta) - f(X;\theta^{\uparrow N}) \|_2,
\end{equation}
where $f(X;\theta)$ is the layer output with parameters $\theta$, and $f(X;\theta^{\uparrow N})$ is the output when neuron $N$ is deactivated (i.e., its parameters are set to zero).

\subsection{Feed-Forward Neurons}

A standard feed-forward network (FFN) layer can be written as:
\begin{equation}
\text{FFN}(X) = \big( \text{SiLU}(XW_{\text{gate}}) \odot (XW_{\text{up}}) \big) W_{\text{down}},
\end{equation}
where $W_{\text{gate}}, W_{\text{up}} \in \mathbb{R}^{d_{\text{model}} \times d_{\text{inter}}}$ and 
$W_{\text{down}} \in \mathbb{R}^{d_{\text{inter}} \times d_{\text{model}}}$. 
The intermediate activation is:
\begin{equation}
H_{\text{act}} = \text{SiLU}(XW_{\text{gate}}) \odot (XW_{\text{up}}) \in \mathbb{R}^{l \times d_{\text{inter}}}.
\end{equation}

Deactivating the $k$-th intermediate neuron corresponds to zeroing out the $k$-th column $H_{\text{act}}[:,k]$, which is equivalent to removing the $k$-th feature before multiplication with $W_{\text{down}}$. 
The resulting change in output is:
\begin{equation}
\Delta Y_{\text{FFN},k} = H_{\text{act}}[:,k] \cdot (W_{\text{down}})_{k,:}.
\end{equation}

Thus, the impact of neuron $k$ is the squared L2 norm:
\begin{equation}
\| \Delta Y_{\text{FFN},k} \|^2 = \| H_{\text{act}}[:,k] (W_{\text{down}})_{k,:} \|^2. \tag{8}
\end{equation}

This can be computed in parallel for all $k \in \{1, \ldots, d_{\text{inter}}\}$.

\subsection{Self-Attention Neurons}

For a single-head self-attention layer:
\begin{equation}
Y_{\text{Attn}} = \text{Softmax}\!\left(\frac{QK^\top}{\sqrt{d_k}}\right) V,
\end{equation}
with $Q = XW_Q$, $K = XW_K$, $V = XW_V$. 
The attention matrix is $A = \text{Softmax}(QK^\top / \sqrt{d_k})$, yielding output $Y_{\text{Attn}} = AV$.

\subsubsection*{(a) Value Neurons ($W_V$)}

A neuron defined by column $k$ of $W_V$ sets $V[:,k]$ to zero when deactivated. 
The change in output is:
\begin{equation}
\Delta Y^{(V)}_{\text{Attn},k} = A V[:,k],
\end{equation}
and the impact is:
\begin{equation}
\| \Delta Y^{(V)}_{\text{Attn},k} \|^2 = \| A V[:,k] \|^2. \tag{10}
\end{equation}

\subsubsection*{(b) Query Neurons ($W_Q$)}

Deactivating the $k$-th column of $W_Q$ zeroes $Q[:,k]$, which alters the unnormalized attention scores:
\begin{equation}
\Delta S_{\text{raw},k} = \frac{Q[:,k] (K[:,k])^\top}{\sqrt{d_k}}.
\end{equation}

This modifies the attention matrix from $A_{\text{orig}} = \text{softmax}(S_{\text{raw}})$ 
to $A^{\uparrow N_{Q,k}} = \text{softmax}(S_{\text{raw}} - \Delta S_{\text{raw},k})$. 
The resulting change in output is:
\begin{equation}
\Delta Y^{(Q)}_{\text{Attn},k} = (A_{\text{orig}} - A^{\uparrow N_{Q,k}}) V,
\end{equation}
and its impact is:
\begin{equation}
\| \Delta Y^{(Q)}_{\text{Attn},k} \|^2. \tag{11}
\end{equation}

\subsubsection*{(c) Key Neurons ($W_K$)}

The effect of deactivating a key neuron is \emph{symmetric} to that of query neurons, 
since the change term
\begin{equation}
\Delta S_{\text{raw},k} = \frac{Q[:,k] (K[:,k])^\top}{\sqrt{d_k}}
\end{equation}
captures the interaction between query and key neurons along dimension $k$. 
Thus, query and key neurons are dual roles in shaping the attention scores.

\subsection{Equivalence of Query and Value Impacts}

Although the mechanics differ---query/key neurons alter the \emph{attention weights}, while value neurons alter the \emph{content propagated}---their impact is functionally equivalent in our parallel detection framework:
\begin{itemize}
    \item Both are measured as L2 changes in the attention output $Y_{\text{Attn}}$.
    \item A query neuron modifies the distribution $A$ over tokens, indirectly scaling the contribution of values.
    \item A value neuron directly zeroes out one content column, effectively applying a structured mask on $V$.
\end{itemize}

Therefore, in practice, \texttt{attn.q} and \texttt{attn.v} neurons are comparable impact carriers, 
differing only in whether they modulate weights (Q) or content (V).
\section{Theory and Proofs for Section~\ref{sec:theory}}
\label{app:proof}

\textbf{Setup.}
We compare \textsc{SNRF} (Eq.~\ref{eq:update}) with linear merging:
\begin{equation}
\begin{aligned}
\widetilde{W}^{\lin} &= W^{\tgt}
 + \beta\,\Delta,\\[2pt]
\Delta &= W^{\src} - W^{\tgt}.
\end{aligned}
\label{eq:lin}
\end{equation}

For theory, we suppress $(\ell,p)$ and work with a single block $W$.
Let $S$ denote the shared-inference coordinates
(Sec.~\ref{sec:method}), with projectors $P_S$ and $P_{\perp}$.
We decompose the update as
\[
\Delta_S := P_S \Delta,
\qquad
\Delta_{\perp} := P_{\perp} \Delta.
\]

\textsc{SNRF} applies a masked rank-$r$ update inside $S$:
\begin{equation}
\begin{aligned}
\widetilde{W}^{\snrf}
 &= W^{\tgt}
 + \beta\,P_S(\Delta^{(r)}),\\[2pt]
\Delta^{\snrf}
 &= \Delta^{(r)}_S,\\[2pt]
\Delta^{\lin}
 &= \Delta_S + \Delta_{\perp},
\end{aligned}
\end{equation}
where $\Delta^{(r)}_S$ is the rank-$r$ truncated SVD of $\Delta_S$.

\textbf{Main result.}
For sufficiently small~$\beta$,
\begin{equation}
\begin{aligned}
\Delta \mathcal{L}_{\lin}(\beta)
 - \Delta \mathcal{L}_{\snrf}(\beta)
\;\ge\;&
\underbrace{
\frac{\beta^2}{2}\,\mu_{\perp}
 \bigl\|\Delta_{\perp}\bigr\|_F^2
}_{\text{curvature in } S^{\perp}\text{ (lin only)}}
\\[6pt]
&\quad
-
\underbrace{
\frac{\beta^2}{2}\,\mu_{S}
 \bigl\|\Delta_S - \Delta^{(r)}_S\bigr\|_F^2
}_{\text{low-rank truncation (SNRF only)}}
\\[6pt]
&\quad
-
\underbrace{
\beta\,c\,
 \bigl\|P_S g\bigr\|_F\,
 \bigl\|\Delta\bigr\|_F
}_{\text{leakage}}
+ O(\beta^3),
\end{aligned}
\label{eq:gap}
\end{equation}
where $c=\varepsilon(1+\eta)$ is small.

Whenever
\begin{equation}
\mu_{\perp}\,
\bigl\|\Delta_{\perp}\bigr\|_F^2
\;>\;
\mu_S\,
\bigl\|\Delta_S - \Delta^{(r)}_S\bigr\|_F^2
\;+\;
c\,\beta^{-1}\,
 \bigl\|P_S g\bigr\|_F\,
 \bigl\|\Delta\bigr\|_F,
\label{eq:cond}
\end{equation}
we obtain
$\Delta\mathcal{L}_{\snrf}(\beta)
 < \Delta\mathcal{L}_{\lin}(\beta)$.

Thus, for sufficiently small $\beta$, linear merging incurs a strictly
higher loss due to movement in $S^{\perp}$, where curvature is large.
\textsc{SNRF} masks~$S^{\perp}$ and pays only the in-$S$ truncation
error---which vanishes whenever
$r=\mathrm{rank}(\Delta_S)$ and is minimized otherwise by the truncated
SVD.

\paragraph{Assumptions (A1-A4).}
\begin{itemize}
  \item[(i)] Gradient concentrates in $S$:
  $\|P_{\perp}g\|_F \le \varepsilon\,\|P_S g\|_F$.
  
  \item[(ii)] Curvature gap:
  $H \succeq \mu_S P_S + \mu_{\perp} P_{\perp}$ with 
  $0 < \mu_S \le \mu_{\perp}$.
  
  \item[(iii)] First-order signal aligns with $S$:
  $\langle P_{\perp} g,\Delta_{\perp}\rangle 
  \ge -\eta\,\|P_{\perp}g\|_F\,\|\Delta_{\perp}\|_F$.
  
  \item[(iv)] $\Delta^{(r)}_S$ is the best rank-$r$ approximation 
  of $\Delta_S$ (Eckart-Young-Mirsky).
\end{itemize}

\paragraph{Taylor expansions.}
At $W^{\tgt}$, let $g=\nabla\mathcal{L}(W)$ and $H=\nabla^2\mathcal{L}(W)$:
\begin{align}
\Delta\mathcal{L}_{\lin}(\beta)
&=\beta\,\langle g,\Delta\rangle 
 + \tfrac{\beta^2}{2}\,\langle \Delta,H\Delta\rangle 
 + O(\beta^3),\\
\Delta\mathcal{L}_{\snrf}(\beta)
&=\beta\,\langle g,\Delta^{(r)}_S\rangle
 + \tfrac{\beta^2}{2}\,\langle \Delta^{(r)}_S,H\Delta^{(r)}_S\rangle
 + O(\beta^3).
\end{align}

\paragraph{Quadratic terms via curvature gap.}
Decompose $\Delta=\Delta_S+\Delta_{\perp}$.  
Using $H\succeq \mu_S P_S + \mu_{\perp} P_{\perp}$,
\[
\langle \Delta,H\Delta\rangle 
\ge \mu_S \|\Delta_S\|_F^2 + \mu_{\perp}\|\Delta_{\perp}\|_F^2,
\qquad
\langle \Delta^{(r)}_S,H\Delta^{(r)}_S\rangle 
\ge \mu_S \|\Delta^{(r)}_S\|_F^2.
\]
Hence,
\[
\langle \Delta,H\Delta\rangle
-
\langle \Delta^{(r)}_S,H\Delta^{(r)}_S\rangle
\ge
\mu_{\perp}\|\Delta_{\perp}\|_F^2
-
\mu_S\|\Delta_S-\Delta^{(r)}_S\|_F^2.
\]

\paragraph{First-order difference and leakage.}
\[
\langle g,\Delta\rangle-\langle g,\Delta^{(r)}_S\rangle
=
\langle P_S g,\Delta_S-\Delta^{(r)}_S\rangle
+
\langle P_{\perp} g,\Delta_{\perp}\rangle.
\]
By Cauchy-Schwarz and (A1)-(A3):
\[
\langle P_S g,\Delta_S-\Delta^{(r)}_S\rangle
\le
\|P_S g\|_F\,\|\Delta_S-\Delta^{(r)}_S\|_F,
\]
\[
\langle P_{\perp} g,\Delta_{\perp}\rangle
\ge 
-\eta\,\|P_{\perp} g\|_F\,\|\Delta_{\perp}\|_F
\ge 
-\eta\varepsilon\,\|P_S g\|_F\,\|\Delta\|_F.
\]

\paragraph{Low-rank optimality in $S$.}
By Eckart-Young-Mirsky,
$\Delta^{(r)}_S$ minimizes 
$\|\Delta_S-\widehat{\Delta}\|_F$ 
over all rank-$r$ matrices $\widehat{\Delta}$.

\paragraph{Collecting terms.}
Combining the bounds above with the Taylor expansion yields Eq.~\ref{eq:gap}.  
Condition~\ref{eq:cond} ensures that the positive quadratic advantage 
from masking $S^{\perp}$ dominates both truncation and leakage terms, 
establishing $\Delta\mathcal{L}_{\snrf}(\beta)<\Delta\mathcal{L}_{\lin}(\beta)$ 
for sufficiently small~$\beta$.
\hfill$\square$

 \section{Experimental details for Findings OF Shared Neurons Basis}

 \subsection{Finding 1}
 \label{app:overlap}
\paragraph{Strong Overlap.}  

\begin{figure*}[t]
    \centering
    \begin{minipage}[t]{0.3\linewidth}
    \centering
    \includegraphics[width=\linewidth]{figs/venn_qwenvl.png}
    \subcaption{Qwen2.5-VL-7B vs Qwen2.5-Math-7B}\label{fig:venn-qwenvl}
  \end{minipage}
      \begin{minipage}[t]{0.3\linewidth}
    \centering
    \includegraphics[width=\linewidth]{figs/venn_intern.png}
    \subcaption{Intern2.5-VL-4B vs Qwen2.5-GRPO-3B}\label{fig:venn-intern}
  \end{minipage}
    \begin{minipage}[t]{0.3\linewidth}
    \centering
    \includegraphics[width=\linewidth]{figs/venn_idefics.png}
    \subcaption{Idefics3 vs Deepseek-LLaMA3-8B}\label{fig:venn-idefics}
  \end{minipage}
      \begin{minipage}[t]{0.3\linewidth}
    \centering
    \includegraphics[width=\linewidth]{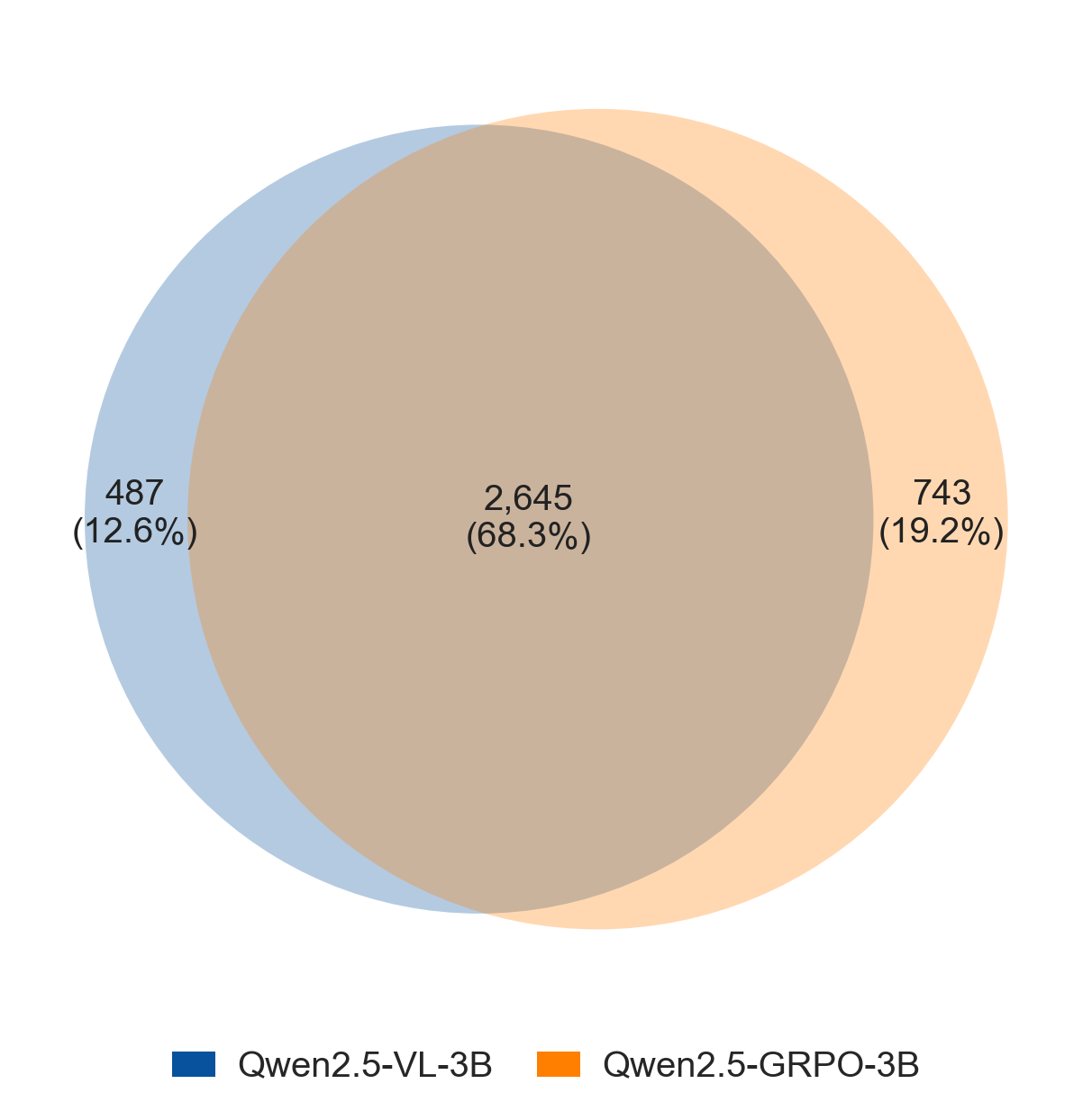}
    \subcaption{Qwen2.5-VL-3B vs Qwen2.5-GRPO-3B}\label{fig:venn-qwenvl_3b}
  \end{minipage}
        \begin{minipage}[t]{0.3\linewidth}
    \centering
    \includegraphics[width=\linewidth]{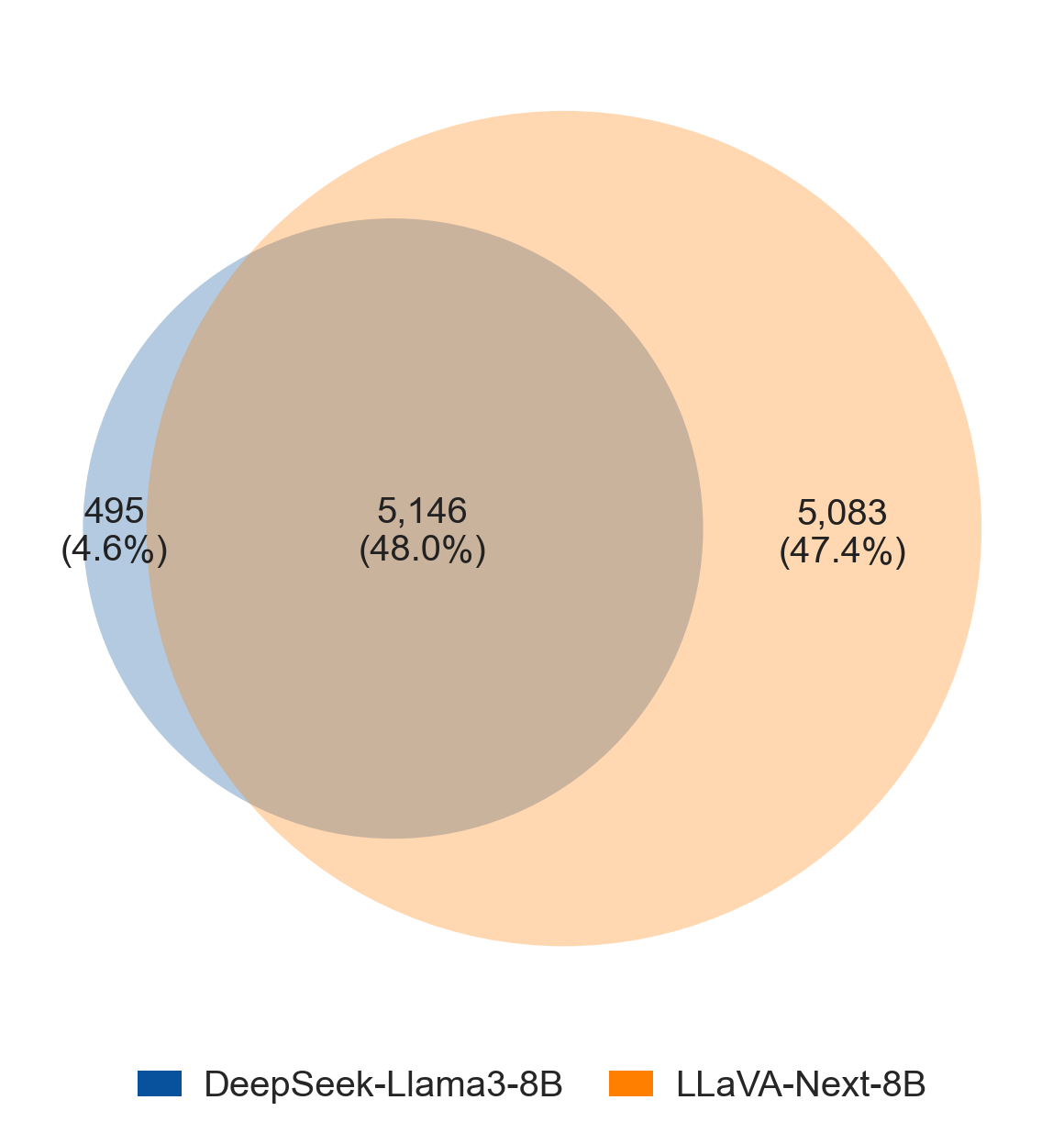}
    \subcaption{LLaVA-Next-8B vs Deepseek-LLaMA3-8B}\label{fig:venn-llavanext}
  \end{minipage}
          \begin{minipage}[t]{0.3\linewidth}
    \centering
    \includegraphics[width=\linewidth]{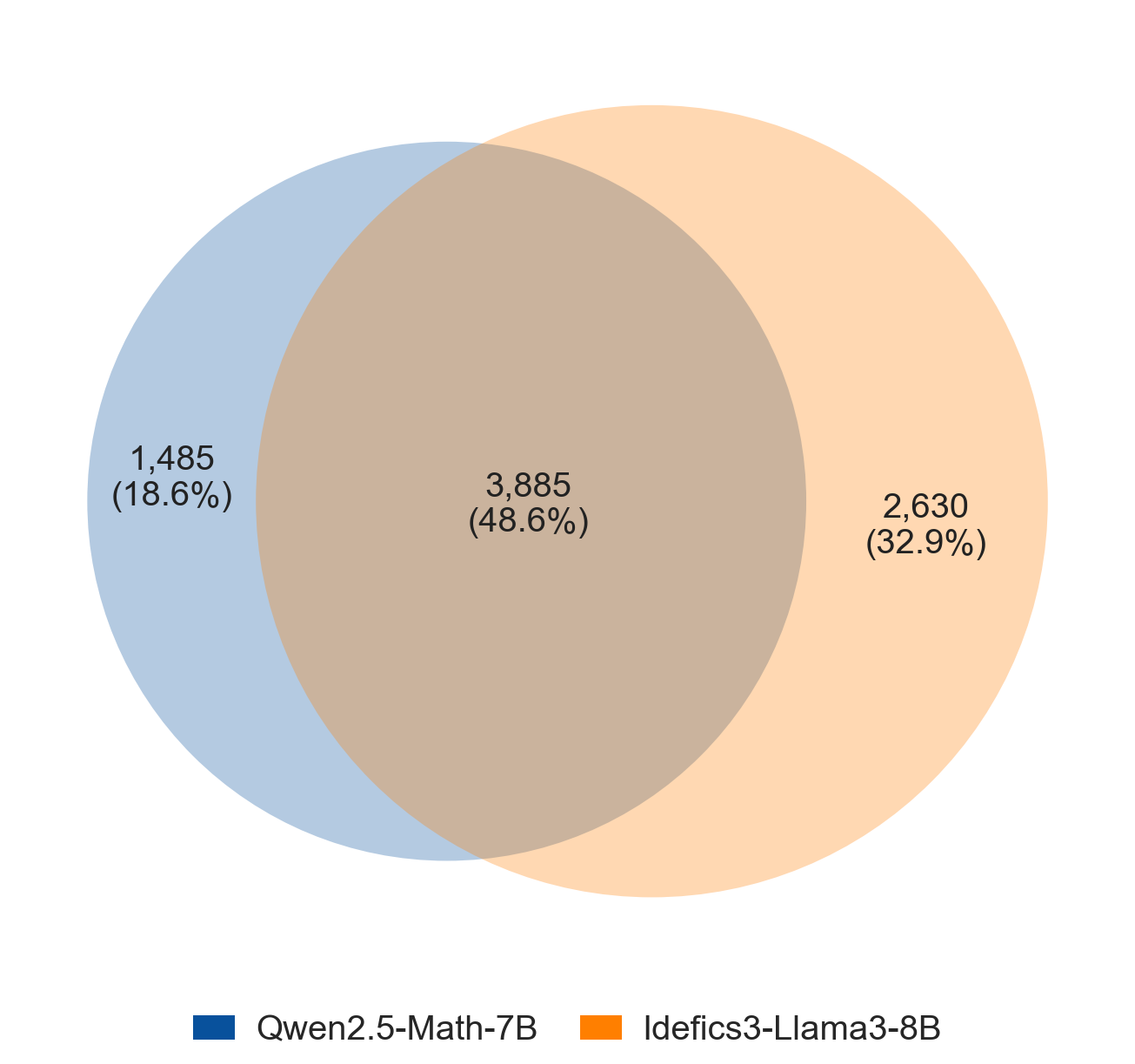}
    \subcaption{Idefics3 vs Qwen2.5-Math-7B}\label{fig:venn-ide_qwenmath}
  \end{minipage}
  \caption{Overlap of activated neurons across different LVLM families.}
  \label{fig:global-venn}
\end{figure*}

\begin{figure*}[h]
    \centering
    \includegraphics[width=0.88\linewidth]{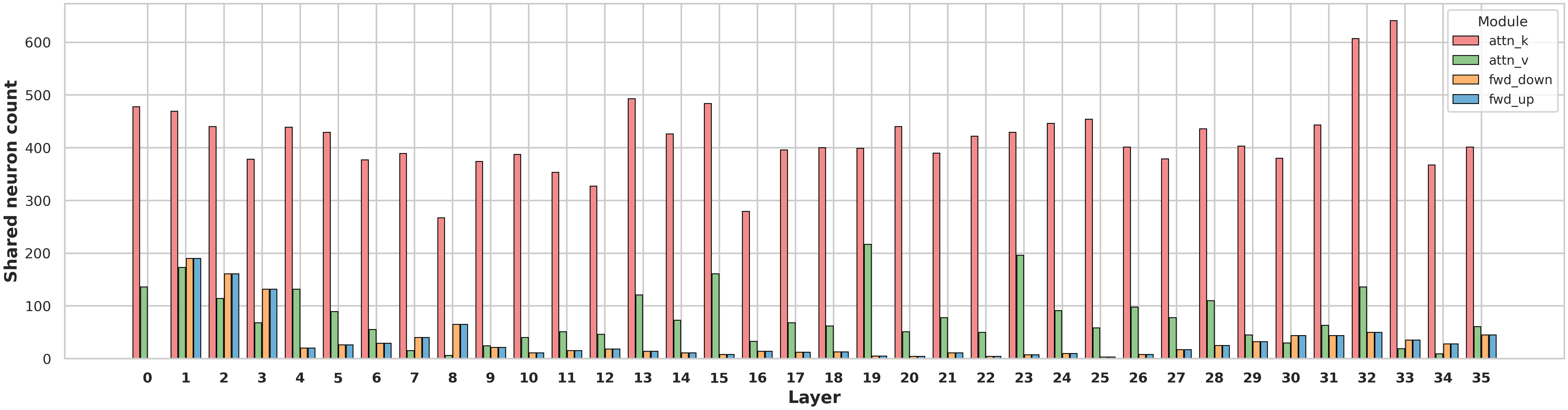}
    \caption{Distribution of shared neurons across layers and modules in Intern2.5-VL-4B and Qwen2.5-GRPO-3B.}
    \label{fig:shared-layer-module_app1}
\end{figure*}

\begin{figure*}[h]
    \centering
    \includegraphics[width=0.88\linewidth]{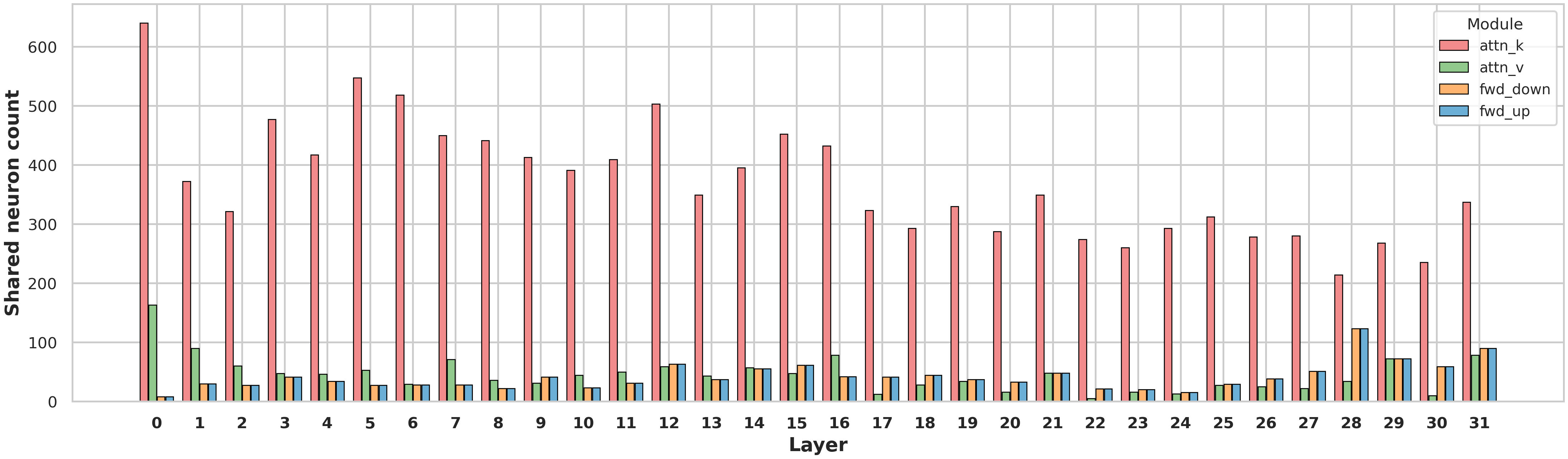}
    \caption{Distribution of shared neurons across layers and modules in Idefics3-LLaMA-8B and Deepseek-LLaMA3-8B.}
    \label{fig:shared-layer-module_app2}
\end{figure*}

Across the five representative model pairs shown in Figure~\ref{fig:global-venn}, we consistently observe substantial overlap in inference-related neurons: Qwen2.5-VL-7B vs.\ Qwen2.5-Math-7B share 4{,}703 neurons (74.5\% of the union) with only 667 (10.6\%) and 942 (14.9\%) remaining model-specific; Intern2.5-VL-4B vs.\ Qwen2.5-GRPO-3B share 2{,}967 (71.3\%); Idefics3-8B vs.\ Deepseek-LLaMA3-8B share 5{,}069 (71.5\%); Qwen2.5-VL-3B vs.\ Qwen2.5-GRPO-3B share 2{,}645 (68.3\%); and even the cross-backbone pair LLaVA-Next-8B vs.\ Deepseek-LLaMA3-8B retains 5{,}146 (48.0\%) shared neurons. Layer-module inspection (Figure~\ref{fig:shared-layer-module_app1} and~\ref{fig:shared-layer-module_app2}) shows that these shared neurons concentrate in \texttt{attn.k} with secondary presence in \texttt{attn.v}, forming two prominent clusters in early layers (before layer~6) and late layers (after layer~20), consistent with roles in generic semantic parsing and high-level inference. Notably, strong overlaps also occur between models without identical language backbones, pointing to universal inference units that transcend architecture and training objectives. Causal ablation confirms their functional importance: zeroing parameters exclusively associated with the shared set ($\mathcal{N}_{\text{shared}}$) leads to a much larger drop in inference accuracy than randomly ablating an equal number of neurons, demonstrating that these shared neurons form a core inference substrate reused across modalities and model families.
Additionally, as shown in Figure~\ref{fig:global-venn}, we observe that even models of different types share a substantial number of neurons.

\paragraph{Where the shared neurons live.}
Building on the main-text observations for Qwen2.5-VL-7B and Qwen2.5-Math-7B, we further examine how shared neurons distribute across layers and modules in other model pairs.
As shown in Figure~\ref{fig:shared-layer-module}, the shared neurons in \textbf{Intern2.5-VL-4B} and \textbf{Qwen2.5-GRPO-3B} are predominantly concentrated in the \texttt{attn.k} matrices across almost all layers, with a consistent but smaller secondary presence in \texttt{attn.v}.
Only a handful of early layers (e.g., layers 1--3) and a few middle or late layers display modest activity in the \texttt{fwd\_down} or \texttt{fwd\_up} modules.
Notably, Intern2.5-VL-4B exhibits an additional rise in shared-neuron counts in the upper layers (around layer 30 and beyond), indicating a stronger reliance on key-vector sharing during deep decoding,
whereas Qwen2.5-GRPO-3B shows relatively more \texttt{attn.v} sharing in the earliest layers, suggesting richer value-vector interactions during initial feature capture.

A similar pattern appears for \textbf{Idefics3-LLaMA-8B} and \textbf{Deepseek-LLaMA3-8B}.
Here, \texttt{attn.k} again dominates the shared-neuron distribution and \texttt{attn.v} provides a secondary contribution, while \texttt{fwd\_down} and \texttt{fwd\_up} remain consistently minor with only localized peaks in the very first or final layers.
Unlike the Qwen family, however, these two models show a relatively flat, plateau-like distribution through the middle layers (approximately layers 5--15) and a sharp spike in the very first one or two layers, pointing to a particularly heavy demand for shared neurons during the initial embedding-fusion stage.

Overall, despite architectural differences, all examined models exhibit a global regularity:
shared neurons overwhelmingly reside in \texttt{attn.k}, with \texttt{attn.v} consistently secondary.
This pattern underscores the central role of attention key vectors in cross-model knowledge sharing and transfer.


 \subsection{Fing 2}
 \label{app:finding2}
  
\paragraph{More experiments on deactivation experiments.}

We evaluate the \emph{causal} role of shared neurons ($\mathcal{N}_{\text{shared}}$; Eq.~\ref{eq:ident-3}) by  
(i) \textbf{Deact}—zeroing the parameters that produce/consume only neurons in $\mathcal{N}_{\text{shared}}$; and  
(ii) \textbf{Random Deact}—ablating the \emph{same number} of neurons sampled at random from the same layer-module budget.

Table~\ref{tab:finding2_full} summarizes the results across \textsc{MathVista} (CoT/Format/Solution), MME, POPE, ScienceQA, MMMU and MMMU-Pro (V).
Across three LVLMs, deactivating the shared neurons collapses performance to \textbf{0.0} on every metric,
whereas random ablations of equal size lead to a moderate drop only.
This pattern holds consistently, demonstrating that $\mathcal{N}_{\text{shared}}$ are both
\emph{necessary} and \emph{specific} for multi-step inference.

\begin{table*}[t]
\centering
\small
\resizebox{\textwidth}{!}{
\begin{tabular}{lcccccccc}
\hline
\textbf{Model} & \textbf{MathVista-Cot} & \textbf{Format} & \textbf{Solution} & \textbf{MME} & \textbf{POPE} & \textbf{SCIQA} & \textbf{MMMU} & \textbf{MMMU-Pro (V)}\\
\hline
\multicolumn{9}{c}{\emph{Base}}\\
\hline
QwenVL-7B      & 48.8 & 68.8 & 37.5 & 1681 & 86.2 & 54.5 & 0.503 & 0.116\\
QwenVL-3B      & 52.6 & 61.6 & 50.8 & 1535 & 87.2 & 52.5 & 0.461 & 0.018\\
Intern2.5-VL-4B    & 60.4 & 65.1 & 61.6 & 1670 & 90.8 & 97.4 & 0.491 & 0.000\\
\hline
\multicolumn{9}{c}{\emph{Deact (shared)}}\\
\hline
QwenVL-7B      & 0.0 & 0.0 & 0.0 & 0 & 0.0 & 0.0 & 0.000 & 0.000\\
QwenVL-3B      & 0.0 & 0.0 & 0.0 & 0 & 0.0 & 0.0 & 0.000 & 0.000\\
Intern2.5-VL-4B    & 0.0 & 0.0 & 0.0 & 0 & 0.0 & 0.0 & 0.000 & 0.000\\
\hline
\multicolumn{9}{c}{\emph{Random Deact}}\\
\hline
QwenVL-7B      & 36.4 & 47.2 & 32.6 & 1510 & 75.9 & 45.6 & 0.482 & 0.081\\
QwenVL-3B      & 32.6 & 32.1 & 32.1 & 1242 & 74.5 & 34.1 & 0.393 & 0.004\\
Intern2.5-VL-4B    & 34.4 & 34.3 & 36.2 & 1353 & 82.2 & 63.4 & 0.272 & 0.000\\
\hline
\end{tabular}
} 
\caption{\textbf{Effect of ablating shared neurons across benchmarks.}  
Removing $\mathcal{N}_{\text{shared}}$ collapses performance to zero,
while random ablations of the same size only moderately reduce scores. 
Higher is better for all metrics.}
\label{tab:finding2_full}
\end{table*}

  \subsection{Fing 3}
 \label{app:finding3}
\begin{figure*}[htbp]
\centering
\small

\begin{subfigure}[b]{0.31\textwidth}
    \includegraphics[width=\textwidth]{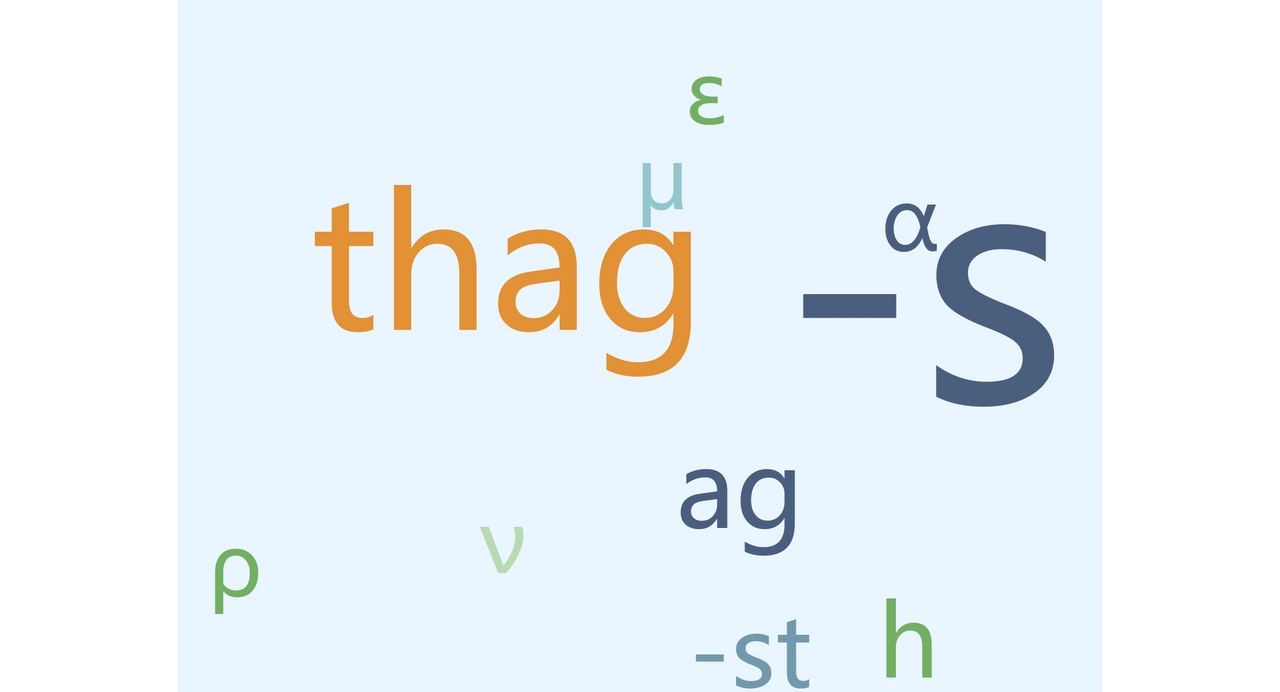}
    \caption{\scriptsize \texttt{ide.fwd.down.L21.N516}}
\end{subfigure}
\hfill
\begin{subfigure}[b]{0.31\textwidth}
    \includegraphics[width=\textwidth]{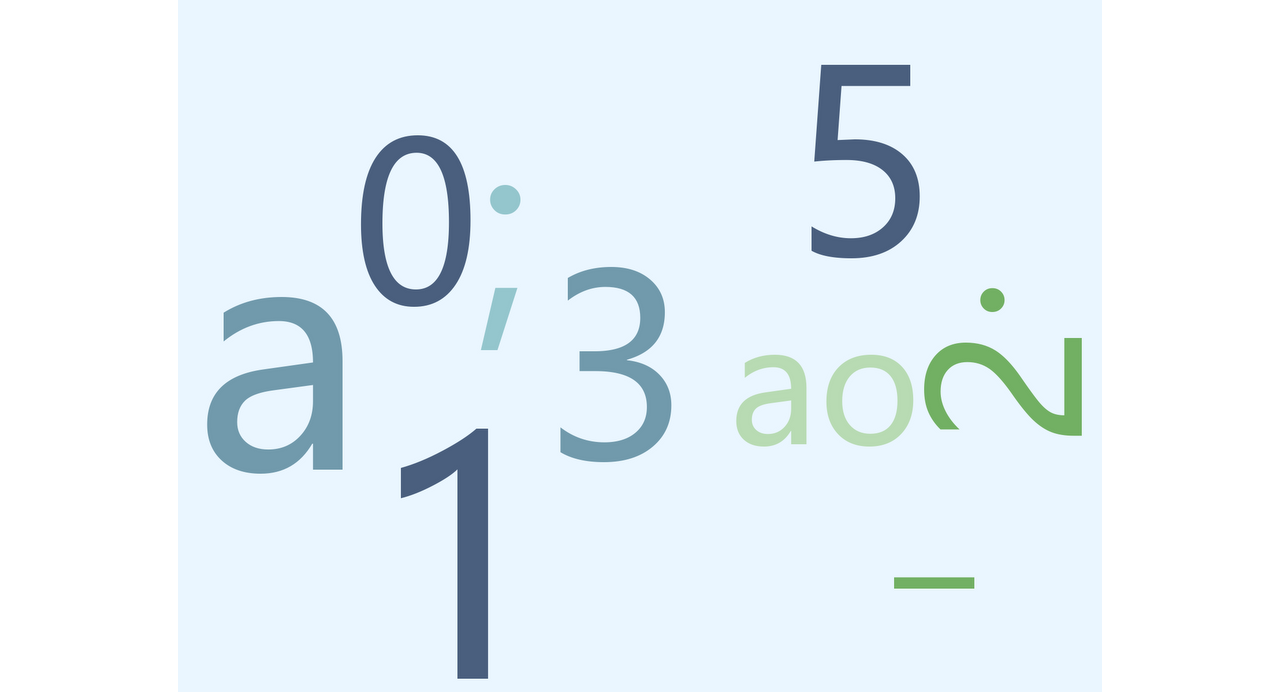}
    \caption{\scriptsize \texttt{ide.fwd.down.L7.N8196}}
\end{subfigure}
\hfill
\begin{subfigure}[b]{0.31\textwidth}
    \includegraphics[width=\textwidth]{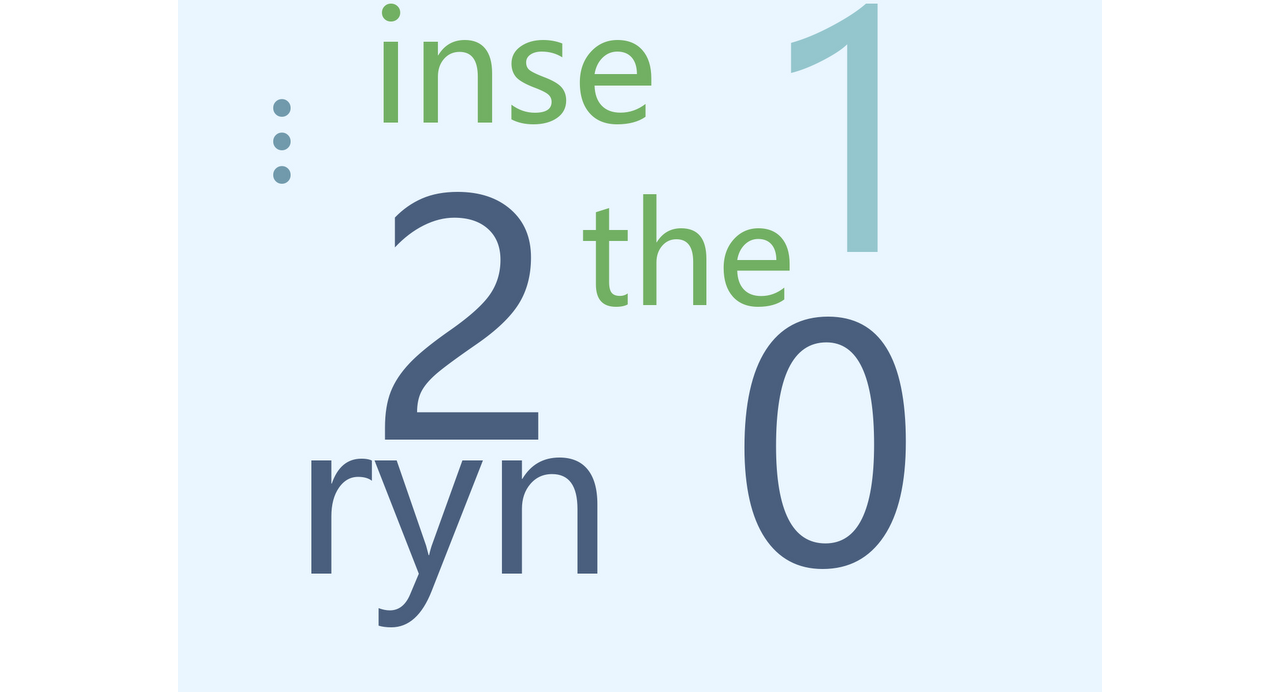}
    \caption{\scriptsize \texttt{LLaMA3.attn.v.L3.N2052}}
\end{subfigure}

\par\medskip 

\begin{subfigure}[b]{0.31\textwidth}
    \includegraphics[width=\textwidth]{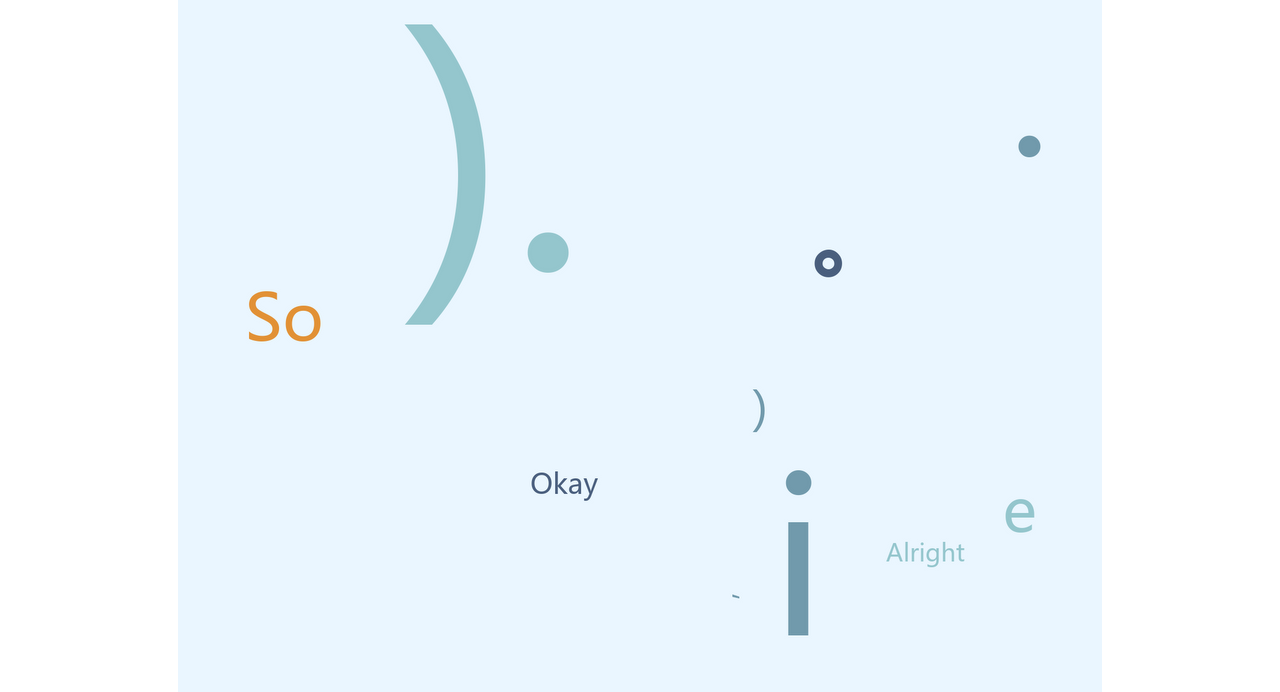}
    \caption{\scriptsize \texttt{LLaMA3.fwd.down.L0.N2378}}
\end{subfigure}
\hfill
\begin{subfigure}[b]{0.31\textwidth}
    \includegraphics[width=\textwidth]{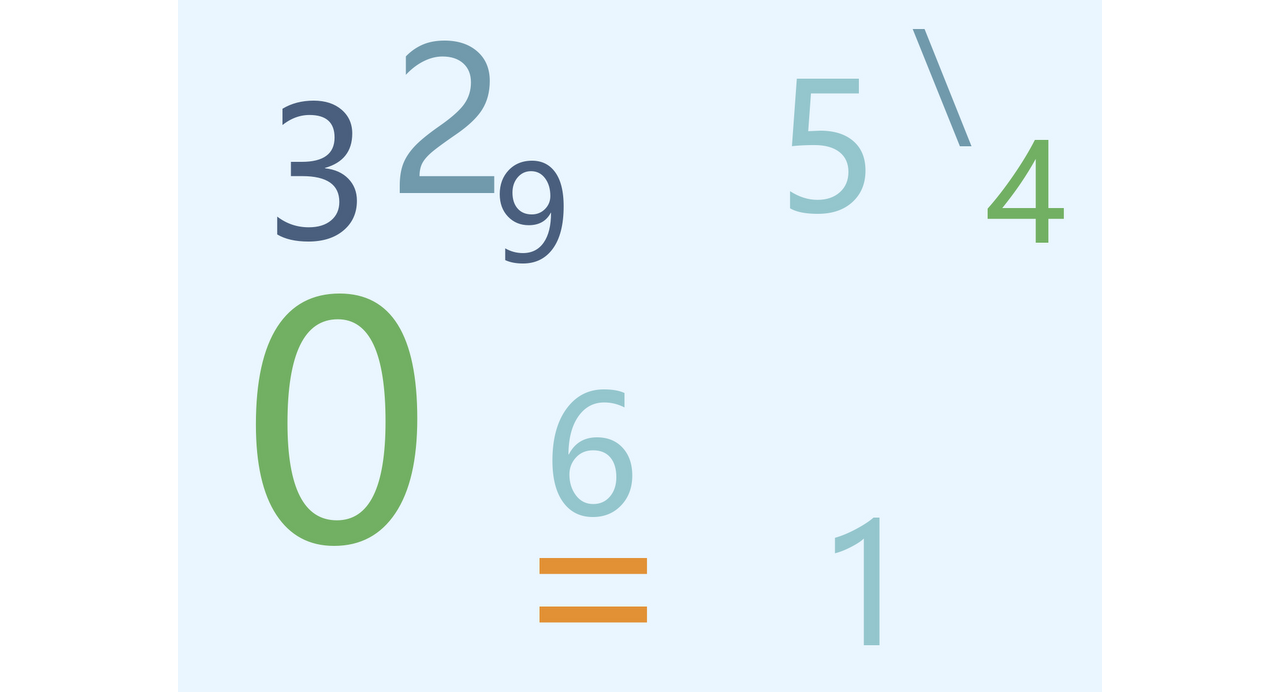}
    \caption{\scriptsize \texttt{intern.fwd.down.L0.N2378}}
\end{subfigure}
\hfill
\begin{subfigure}[b]{0.31\textwidth}
    \includegraphics[width=\textwidth]{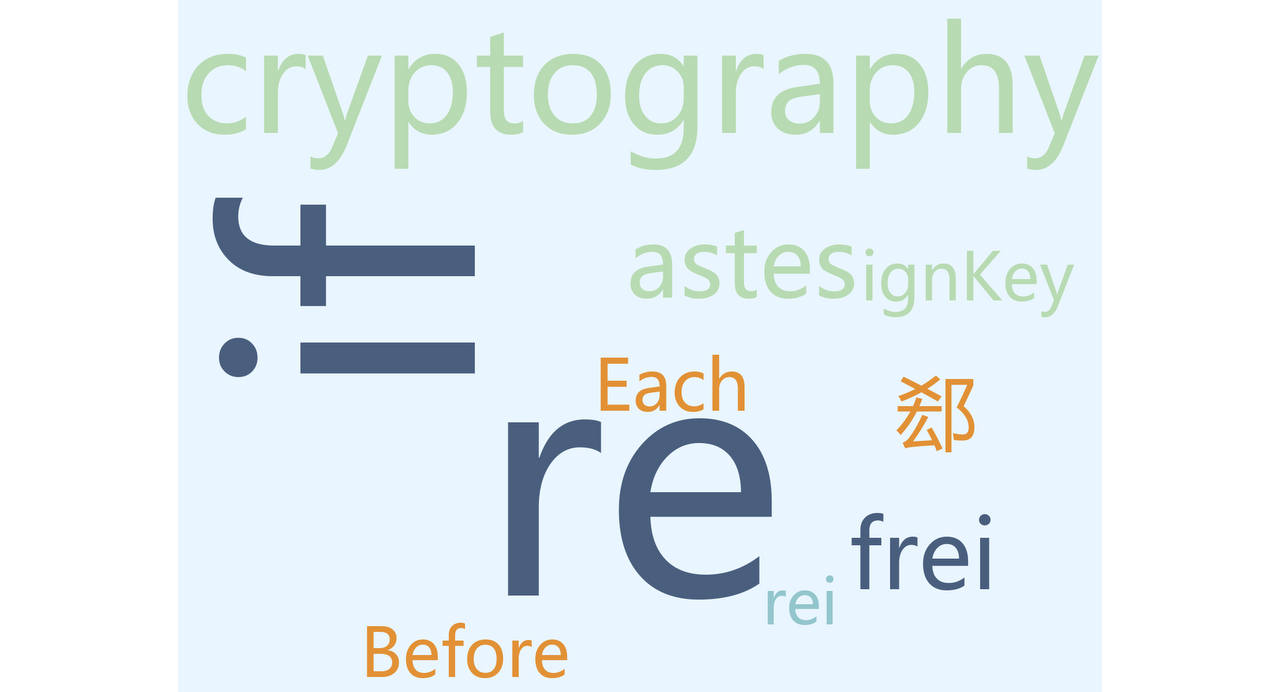}
    \caption{\scriptsize \texttt{qwen3b.attn.v.L0.N2378}}
\end{subfigure}

\caption{Word clouds of token concepts for different neurons.}
\label{fig:wordclouds}
\end{figure*}

 \paragraph{More Cases.} Using our amplification intervention, from Figure \ref{fig:wordclouds} presents six cases. In Idefics3’s feed-forward down-projection, \texttt{ide.fwd.down.L21.N516} (a) favors subword fragments composed of letters and symbols (mixed case letters, hyphens, Greek marks), indicating sensitivity to morphological pieces; \texttt{ide.fwd.down.L7.N8196} (b) instead concentrates on brackets, digits, and simple operators, reflecting selectivity for equation layout. In LLaMA3, the attention value channel \texttt{LLaMA3.attn.v.L3.N2052} (c) repeatedly activates on high-frequency function words and number words (e.g., ``the,'' ordinals/cardinals), suggesting a role in anchoring syntax and count-related context, while the feed-forward down-projection \texttt{LLaMA3.fwd.down.L0.N2378} (d) responds more to sentence boundaries and punctuation (right parentheses, periods), encoding formatting cues. Across models, Intern2.5-VL’s \texttt{intern.fwd.down.L0.N2378} (e) shows a strong bias toward numerals and operators, whereas Qwen-3B’s \texttt{qwen3b.attn.v.L0.N2378} (f) highlights topical clusters (e.g.,  cryptography, the ``re-'' prefix). Altogether, these word clouds indicate that, even across models and layers, individual neurons consistently carry sub-domain semantics—digits/operators, typography/punctuation, syntactic anchors, and topical terms—supporting their utility as actionable semantic carriers for structured interpretability and targeted control.

\paragraph{Statistical Analysis} 
We conducted a statistical analysis of the math-related associations within the neurons’ top-10 amplified tokens. 
Using GPT-4o as a classification model, we categorized each token to determine whether it is associated with mathematical concepts.  Let $\mathcal{N}$ denote the set of neurons, and for each neuron $i \in \mathcal{N}$, let 
$T_i = \{t_{i1}, t_{i2}, \ldots, t_{iK}\}$ represent the top-$K$ amplified tokens, with $K=10$ in our analysis. 
Define the indicator function:
\[
\mathbf{1}_{\text{math}}(t) = 
\begin{cases}
1, & \text{if token $t$ is classified as math-related}, \\
0, & \text{otherwise}.
\end{cases}
\]

\begin{figure*}[htbp]
    \centering
    \includegraphics[width=0.85\linewidth]{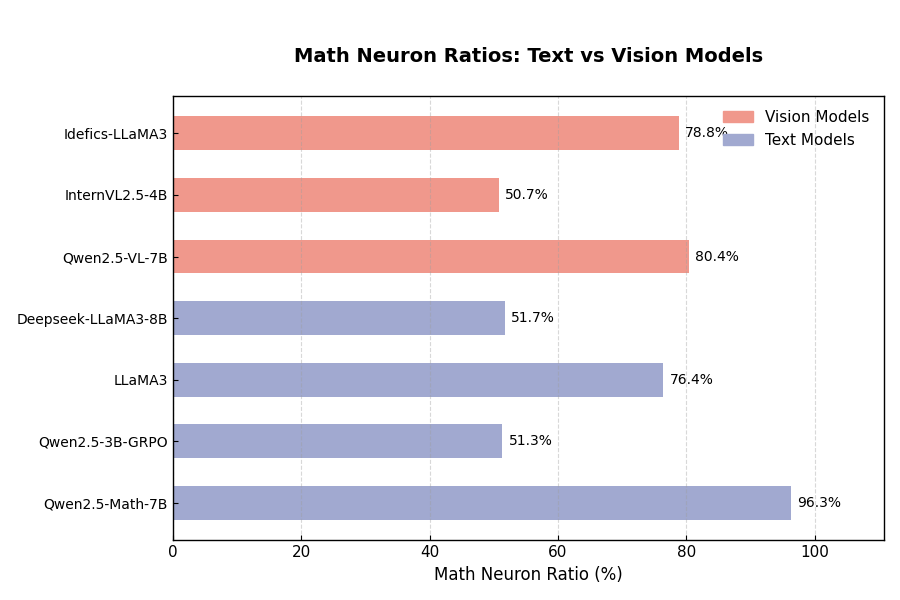}
    \caption{Math neuron ratios across text and vision models. 
    Vision models (red) and text models (blue) are compared by their proportion of math-related neurons, 
    showing that a substantial fraction of shared neurons are activated by mathematical concepts.}
    \label{fig:stat_math_neuron}
\end{figure*}

As shown in Figure~\ref{fig:stat_math_neuron}, the math neuron ratios differ across text and vision models. Text models (blue) generally contain a higher proportion of math-related neurons, with Qwen2.5-Math-7B reaching 96.3\% and LLaMA3 at 76.4\%. Vision-language models (red) also exhibit strong math sensitivity, for example Qwen2.5-VL-7B at 80.4\% and Idefics-LLaMA3 at 78.8\%. These results indicate that both text and vision models encode substantial mathematical knowledge, and that inference capabilities are closely tied to neurons sensitive to mathematical tokens.

These results show that a significant proportion of shared neurons encode mathematical knowledge.

 \section{Details about experiments}
\label{app:exp-details}
\subsection{Exact model specs and checkpoints}
\label{app:models}

Below are the vision‐language models we evaluate, along with their parameter scales, language backbones, paired text models (for parameter‐merging where applicable), and  official checkpoints.

\begin{table*}[ht]
\centering
\resizebox{\textwidth}{!}{%
\begin{tabular}{lllll}
\toprule
\textbf{Model Name} & \textbf{Params} & \textbf{Vision Encoder / Language Backbone} & \textbf{Merged / Text Backbone} & \textbf{Checkpoint} \\
\midrule
Idefics3‐8B‐LLaMA3 & 8B & SigLip vision + LLaMA3 (8B) & DeepSeek-LLaMA3 & \url{HuggingFaceM4/Idefics3-8B-LLaMA3} \\
LLaVA-Next-8B & 8B & (vision encoder as used in LLaVA-NeXT) + LLaMA3 (8B) & DeepSeek-LLaMA3 & \url{lmms-lab/LLaMA3-LLaVA-Next-8B} \\
Qwen2.5-VL-7B & 7B & Qwen2.5 vision + Qwen2.5-VL backbone & Qwen2.5-7B-Math & \url{Qwen/Qwen2.5-VL-7B-Instruct} \\
 Intern2.5-VL-4B & 4B & InternViT vision +  Intern2.5-VL & Qwen2.5-3B-GRPO  & \url{OpenGVLab/InternVL2_5-4B} \\
Qwen2.5-3B-Instruct-Math-GRPO & 3B & — (text-only model) & Qwen2.5-3B-GRPO & \url{Williammsq/Qwen2.5-3B-Instruct-Math-GRPO} \\
\bottomrule
\end{tabular}
} 
\end{table*}

 All models use their authors’ officially released vision encoders. For parameter merging, we follow the published pairing: e.g.\ Qwen2.5-VL-7B is paired with the Qwen2.5-Math text model;  Intern2.5-VL-4B is paired with Qwen3B-GRPO (or equivalent text model) in our merging setup. Model licenses are as released on their model hub pages (e.g.\ Apache-2.0 for most).  


%

\subsection{Details about benchmarks}
\label{app:benchmarks}
We conduct experiments on six representative multimodal benchmarks—MathVista~\citep{lu2023mathvista}, \emph{MME}~\citep{mme}, \emph{POPE}~\citep{pope}, \emph{ScienceQA}~\citep{sciqa}, \emph{MMMU}~\citep{mmmu}, and \emph{MMMU-Pro}~\citep{mmmu_pro}.  
These benchmarks jointly span visual mathematical inference (MathVista), broad perception and cognition (MME), hallucination detection (POPE), multimodal science question answering with explanations (ScienceQA), and college-level multi-discipline understanding and inference (MMMU and MMMU-Pro).  
Together they allow us to measure modality alignment, perception$\to$inference competence, and safety/calibration under diverse conditions.

MathVista provides over six thousand high-quality visual math problems covering functions, geometry, algebra, and real-world quantitative inference. We follow the official split by using the small \texttt{testmini} set for ablations and reporting final results on the hidden \texttt{test} set via the evaluation server. MME evaluates a wide spectrum of perception and inference sub-tasks such as fine-grained recognition, text understanding, logical inference, and commonsense; it provides only a test-style question-answer collection without train/validation subsets and is therefore used in a strict zero/low-shot setting. POPE is designed for hallucination detection by constructing yes/no questions on real-world images with three negative-sampling protocols (\texttt{random}, \texttt{popular}, \texttt{adversarial}); we report all corresponding metrics including accuracy, precision, recall, F1, and the yes-ratio as required by the official script. ScienceQA contains multi-choice science questions enriched with lecture notes and human-written explanations. We do not fine-tune on its training set but use the \texttt{val} split for prompt development and evaluate on the hidden \texttt{test} set, optionally eliciting chain-of-thought inference when explanations are requested.

The original MMMU benchmark covers six major disciplines, more than thirty subjects, and over one hundred and eighty fine-grained sub-fields at college level. It provides \texttt{dev}, \texttt{val}, and hidden \texttt{test} sets and requires submission to an official evaluation server; we use only the public \texttt{dev} and \texttt{val} data for prompt and parameter tuning and report official \texttt{test} scores. MMMU-Pro raises the difficulty with vision-only protocols, stricter filtering of text-answerable questions, and stronger distractor design, offering a more robust evaluation of cross-domain inference.

Inference follows unified prompting. For multiple-choice tasks we instruct the model to output only the letter of the final answer, and for POPE we constrain output to \{\texttt{yes}, \texttt{no}\}. Multi-image questions are fed as ordered lists with minimal positional hints. Unless chain-of-thought or self-consistency is explicitly required, decoding temperature is set to $0$; for CoT we use temperature $0.7$ and majority vote over five samples. Metrics are primarily accuracy for MathVista, ScienceQA, MME, MMMU, and MMMU-Pro, with per-subdomain breakdowns where applicable, while POPE additionally reports precision, recall, F1, and yes-ratio.
Finally, we respect all dataset licenses and use the data strictly for academic research.  
\end{document}

%% file: preamble.tex
\usepackage{subcaption}
\usepackage{algorithm}
\usepackage{amsmath, amssymb}
\usepackage{ifthen}
\usepackage{url}
\usepackage{booktabs}
\usepackage{algpseudocode}
\usepackage[dvipsnames,table]{xcolor}
\usepackage{colortbl}
\usepackage{paracol}
\usepackage{wrapfig}
\usepackage[most]{tcolorbox}
\usepackage{enumitem}
\usepackage{amsthm}
\usepackage{bm}
\usepackage{mathtools}
\usepackage{subcaption}
\usepackage{algorithm}
\usepackage{amsmath, amssymb}
\usepackage{ifthen}
\usepackage{url}
\usepackage{booktabs}
\usepackage{algpseudocode}
\usepackage[dvipsnames,table]{xcolor}
\usepackage{colortbl}
\usepackage{paracol}
\usepackage{wrapfig}
\usepackage[most]{tcolorbox}
\usepackage{enumitem}
\usepackage{amsthm}
\usepackage{bm}
\usepackage{mathtools}